\documentclass[10pt,twocolumn,letterpaper]{article}

\usepackage[pagenumbers]{wacv} % To force page numbers, e.g. for an arXiv version
\definecolor{wacvblue}{rgb}{0.21,0.49,0.74}
\usepackage[pagebackref,breaklinks,colorlinks,allcolors=wacvblue]{hyperref}

\usepackage{etoolbox}
\usepackage{graphicx}
\usepackage{caption}
\usepackage{subcaption} % custom
\usepackage[inkscapelatex=false]{svg} % custom
\usepackage{booktabs}

\makeatletter
\let\old@maketitle\@maketitle
\renewcommand{\@maketitle}{%
  \old@maketitle
  \vspace{-1.5ex}
  \begin{center}
    \includegraphics[scale=0.4]{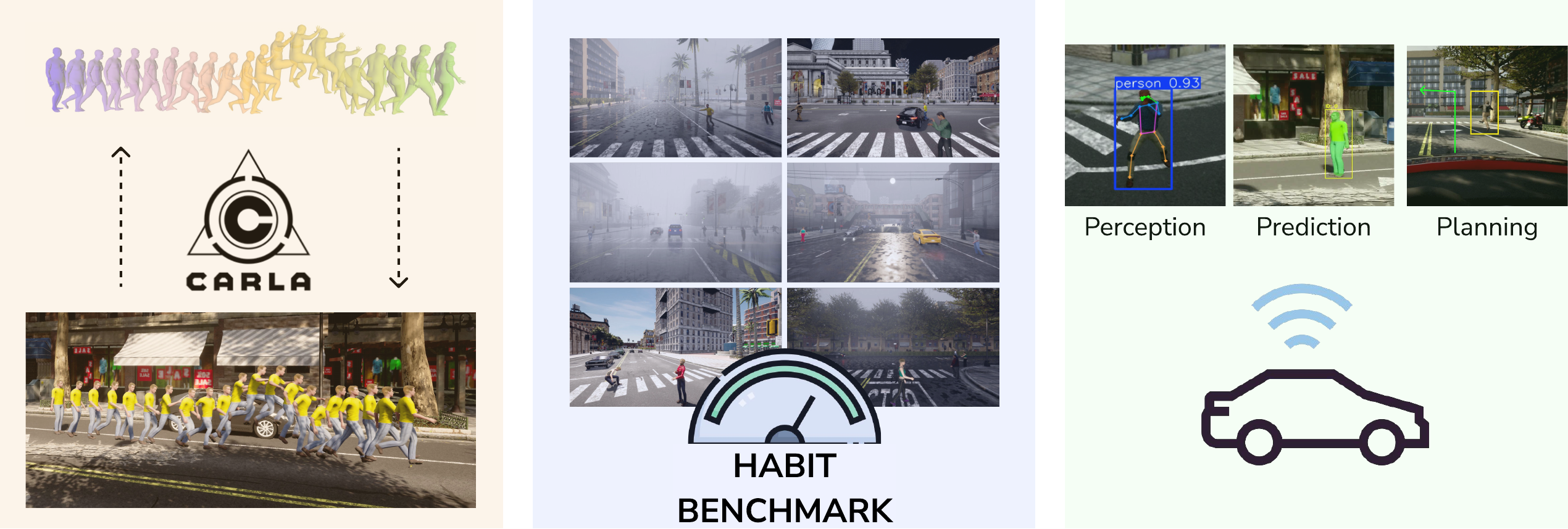} % <== Update path as needed
    \captionof{figure}{HABIT Benchmark: Integrating realistic pedestrian behavior into CARLA for high-fidelity autonomous driving evaluation. Left: retargeting real human motion into simulation. Center: interactive traffic scenarios with diverse agents.   Right: testing perception, prediction, and planning systems under realistic pedestrian dynamics.}
    \label{fig:teaser}
  \end{center}
  \vspace{1ex}
}
\makeatother
\def\wacvPaperID{1565} % *** Enter the WACV Paper ID here
\def\confName{WACV}
\def\confYear{2026}

\title{HABIT: Human Action Benchmark for Interactive Traffic in CARLA}

\author{
Mohan Ramesh\thanks{These authors contributed equally.} \quad
Mark Azer\footnotemark[1] \quad
Fabian B. Flohr \\
\small \textbf{Intelligent Vehicles Lab, Munich University of Applied Sciences}\\
{\tt\small \{mohan.ramesh, mark.azer, fabian.flohr\}@hm.edu}
}

\begin{document}
\maketitle
\begin{abstract}

Current autonomous driving (AD) simulations are critically limited by their inadequate representation of realistic and diverse human behavior, which is essential for ensuring safety and reliability. Existing benchmarks often simplify pedestrian interactions, failing to capture complex, dynamic intentions and varied responses critical for robust system deployment. To overcome this, we introduce HABIT (Human Action Benchmark for Interactive Traffic), a high-fidelity simulation benchmark. HABIT integrates real-world human motion, sourced from mocap and videos, into CARLA~\cite{Dosovitskiy17} (Car Learning to Act, a full autonomous driving simulator) via a modular, extensible, and physically consistent motion retargeting pipeline. From an initial pool of approximately 30,000 retargeted motions, we curate 4,730 traffic-compatible pedestrian motions, standardized in SMPL format for physically consistent trajectories. HABIT seamlessly integrates with CARLA's Leaderboard, enabling automated scenario generation and rigorous agent evaluation. Our safety metrics, including Abbreviated Injury Scale (AIS) and False Positive Braking Rate (FPBR), reveal critical failure modes in state-of-the-art AD agents missed by prior evaluations. Evaluating three state-of-the-art autonomous driving agents, InterFuser, TransFuser, and BEVDriver, demonstrates how HABIT exposes planner weaknesses that remain hidden in scripted simulations.
Despite achieving close or equal to zero collisions per kilometer on the CARLA Leaderboard, the autonomous agents perform notably worse on HABIT, with up to $7.43$ collisions/km and a $12.94$\% AIS 3+ injury risk, and they brake unnecessarily in up to 33\% of cases.
All components are publicly released to support reproducible, pedestrian-aware AI research.

\end{abstract}

% The ABSTRACT is to be in fully justified italicized text, at the top of the left-hand column, below the author and affiliation information.
% Use the word ``Abstract'' as the title, in 12-point Times, boldface type, centered relative to the column, initially capitalized.
% The abstract is to be in 10-point, single-spaced type.
% Leave two blank lines after the Abstract, then begin the main text.
% Look at previous \confName\ abstracts to get a feel for style and length.

%wtt in wacv - https://ieeexplore.ieee.org/stamp/stamp.jsp?tp=&arnumber=10943756    
\section{Introduction}
\label{sec:intro}

Autonomous driving agents face their greatest test in environments shaped by the unpredictable nature of human behavior. Pedestrian dynamics, in particular, introduce a uniquely complex challenge: their movements are rarely deterministic, instead being influenced by implicit social cues, environmental context, and their underlying intentions.
Despite significant advancements in the visual realism of simulation platforms such as CARLA, LGSVL~\cite{rong2020lgsvl}, AirSim~\cite{shah2018airsim}, and DeepDrive~\cite{chen2019deepdriving}, most current benchmarks fall short in capturing the full semantic and kinematic diversity of pedestrian behavior. While tools such as Omniverse~\cite{nvidia_omni_anim_people_2023} offer extensions for populating scenes with animated human actors and enabling basic scripted actions, they still lack support for generating diverse, context-aware trajectories that are truly grounded in real-world pedestrian dynamics. Instead, simulated agents are typically governed by simplistic, handcrafted state machines or deterministic scripts. This results in highly predictable, semantically shallow interactions that ultimately undermine the robustness and relevance of autonomous driving evaluation pipelines.
This behavioral gap leads to models that overfit to synthetic priors, such as the assumption that pedestrians will always cross linearly or yield predictably. Consequently, these models exhibit poor generalization when confronted with ambiguous, rare, or socially nuanced behaviors in the real world. These limitations are especially critical in safety-sensitive scenarios, where accurate forecasting of human intent is essential. The current narrow focus on binary collision metrics in evaluations, which often overlooks the severity of impact, further exacerbates this issue. Without clinically meaningful safety measures, planners may overreact to low-risk encounters while inadvertently underestimating genuinely dangerous ones, thereby distorting our assessment of system competence.

To bridge the significant behavioral gap in autonomous driving simulations, we need to embed realistic human motion at scale, while ensuring both physical realism and contextual consistency. Our solution, HABIT (Human Action Benchmark for Interactive Traffic, see Figure~\ref{fig:teaser}), directly addresses this need.
HABIT is based on a novel pipeline that integrates heterogeneous motion capture and video data into the CARLA simulator, generating highly realistic pedestrian trajectories.
From over 30,000 motion clips, we meticulously curated 4,730 traffic-compatible sequences through a process of semantic filtering and visual validation. 
Unlike static motion clips, HABIT reconstructs temporally coherent trajectories by fusing SMPL~\cite{loper2015smpl} pose sequences with orientation-aligned global trajectory estimation. The resulting motions are smooth, physically grounded, and adaptable to diverse urban environments, faithfully capturing complex behaviors like jaywalking, non-yielding, and hesitant crossing. This rich behavioral diversity is crucial for developing robust autonomous driving systems capable of generalizing to real-world complexities.
We validate HABIT's impact across critical perception, prediction, and planning tasks using state-of-the-art models like YOLOv11-Pose~\cite{khanam2024yolov11}, Segment Anything Model (SAM)~\cite{kirillov2023segment} and InterFuser~\cite{shao2023sinterfuser}. Crucially, HABIT enables us to move beyond simple binary collision metrics by incorporating the Abbreviated Injury Scale (AIS) for a clinically grounded measure of injury severity, allowing for more meaningful safety evaluations.

Our key contributions are threefold.
First, we introduce HABIT, the first CARLA-based benchmark that integrates thousands of semantically curated, real-world pedestrian motions with context-aware behavioral transitions. Our benchmark includes 4,730 high-fidelity traffic-compatible motions, further enhanced with randomized augmentations to increase behavioral diversity.
Second, we develop a modular motion reconstruction pipeline that transforms heterogeneous motion sources, including motion capture and video data, into smooth, physically plausible SMPL-based trajectories. These are aligned globally for temporal coherence and integrated seamlessly into interactive driving simulations.
Third, we propose novel evaluation metrics to assess model behavior beyond binary collision rates. We introduce a false-positive braking metric, revealing conservative failure modes in state-of-the-art planners, and a severity-aware collision assessment using the Abbreviated Injury Scale (AIS), enabling clinically meaningful evaluations of driving risk. Using these metrics, we evaluate InterFuser, TransFuser, and BEVDriver, showing that HABIT reveals planner-specific weaknesses and safety-conservatism trade-offs hidden in scripted benchmarks.
We release all our 30,000 retargeted pedestrian motions, along with tools for motion processing, scenario generation, and evaluation to support reproducible and scalable research.

%\begin{itemize}
%\item HABIT benchmark: The first CARLA-based benchmark integrating thousands of semantically curated real‑world pedestrian motions with context-aware behavior transitions. Our initial release contains $4{,}730$ motions with randomized augmentations to extend motion diversity. Publicly available dataset of over $30{,}000$ retargeted pedestrian motions, along with code for benchmarking, motion processing, and scenario generation to support reproducible research.

%\item Motion reconstruction pipeline: A modular retargeting pipeline converting diverse human motion datasets and video data from any source into physically plausible, SMPL-based trajectories suitable for CARLA integration.

%\item Novel evaluation metrics: Introduction of nuanced, downstream evaluations including testing on the best publicly available model on CARLA leaderboard:
%    \begin{itemize}
%        \item False-positive braking metric showing \textbf{$33\%$} on Habit-Full and \textbf{$63.9\%$} on Habit-Idle variants, showing extreme conservatism.
%        \item Severity-aware collision assessment using the Abbreviated Injury Scale (AIS), we observe MAIS3+ cases of \textbf{$10.96\%$} on Habit-Full and  \textbf{$8.4\%$} on Habit-Idle.
%    \end{itemize}

%\end{itemize}
\section{Related Works}
\label{sec:works}

\begin{figure*}[t]
    \centering
    \begin{minipage}[c][6cm][c]{0.95\textwidth}
        \centering
        \includegraphics[width=\linewidth]{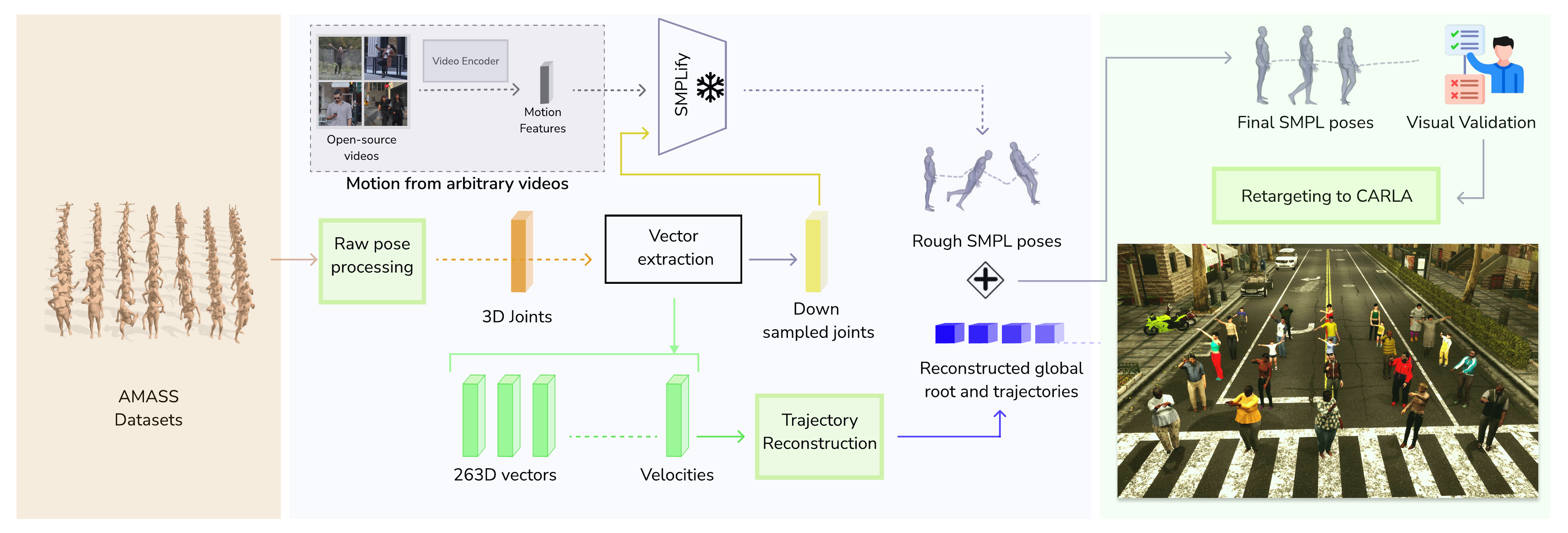}
    \end{minipage}
    \caption{Overview of the proposed motion data processing pipeline. The dotted box presents HABIT'S extensibility and scalability using our video based motion extraction.}
    \label{fig:architecture}
\end{figure*}

\paragraph{Pedestrian Realism in Simulation Environments}
High-fidelity simulation is essential for autonomous driving development and evaluation, yet a persistent reality gap remains between simulated and real human behaviors. Here, we use the term “high-fidelity” to denote simulations that jointly optimize for appearance fidelity (photorealistic rendering), motion fidelity (biomechanically and physically consistent movements), and behavioral fidelity (context-aware, semantically diverse interactions). Platforms like CARLA~\cite{Dosovitskiy17}, LGSVL~\cite{rong2020lgsvl}, AirSim~\cite{shah2018airsim}, and NVIDIA Omniverse (DRIVESim)~\cite{nvidia2021drivesim} offer photorealistic environments and sensor fidelity. However, they still rely heavily on rule-based or deterministic pedestrian behaviors, resulting in semantically shallow and predictable interactions. Although they emphasize addressing both appearance and content gaps, particularly in actor behavior generation, the focus has predominantly been on vehicle traffic and sensor realism, not pedestrian-driven dynamics~\cite{Tan2021SceneGenLT}. Furthermore, Omniverse supports data-driven human motions using USD-based animation blending, though its emphasis remains on high-fidelity rendering rather than pedestrian-interactive behavior. Prior work on pedestrian realism, such as hierarchical behavior models \cite{turn0academia23}, remains sparse and relatively immature. To our knowledge, no existing work integrates real-world pedestrian motion data into a large-scale, context-aware benchmark suitable for pedestrian-aware applications.

To enhance realism, generative and physics-based approaches are gaining traction. Wang~\etal~\cite{Wang2024PacerPlus} introduced PACER+, a reinforcement-learned humanoid controller that follows a trajectory while reproducing motion styles and gestures in a physically plausible manner. This enables diverse behaviors, e.g., evasive movement, turning, or gesturing, grounded in both physics and scene context, however, the work does not scale to large scale interactive 3D simulation environment. Ramesh and Flohr~\cite{10588860} propose Walk-the-Talk a text to motion framework driven by large language models to generate semantically grounded pedestrian motion sequences such as jaywalking, intoxicated pedestrians for integration into a simulated environment. Other systems include the VRU Simulator~\cite{fischer2022vru}, which is designed to evaluate pedestrian and cyclist interactions with vehicles, and GeoSim~\cite{chen2021geosim}, which composites video snippets of real pedestrians into synthetic scenes. These approaches contribute behavioral diversity but often lack environmental grounding or integration into structured simulation and evaluation environments at scale.

% Rock~\etal~\cite{rock2022mocap} conceptualize enhancing pedestrian realism in CARLA by replacing canned animations with motion-captured gestures to achieve more human-like dynamics. In follow-up work, they extend this approach to enable real-time interaction between real pedestrians and simulated vehicles using VR and motion capture, allowing dynamic, human-in-the-loop behavior studies within the CARLA environment.
 %The work remains at the conceptual level, without a concrete implementation or integration into simulation pipelines. Key challenges, such as motion retargeting, environmental grounding, and scenario-driven evaluation—are not addressed, limiting its applicability for scalable or systematic benchmarking.

While these efforts mark significant progress toward realistic human modeling, they remain limited in scalability, contextual diversity, or evaluation integration.

\paragraph{Pedestrian-Aware Prediction and Planning}
Classic models like the Social Force model~\cite{helbing1995social} have been used to simulate pedestrian movements in traffic; while interpretable, these physics-based models require careful parameter tuning and struggle to capture the nuanced interactions of real pedestrians. Current state of research emphasizes integrating pedestrian intent and interaction into planning. Möller~\etal~\cite{moller2025pedestrianaware} combine social force-based motion with risk-aware planning for navigating simulated crowded scenes. Probabilistic approaches further enhance prediction in simulation: Zhang~\etal~\cite{zhang2021improved} model latent intent using Markov Decision Processes, while Dang~\etal~\cite{dang2025dynamic} apply game theory to predict pedestrian actions at unsignalized crossings. Occlusion-aware methods also improve safety realism. Tang~\etal~\cite{tang2021phantom} use phantom agents based on map semantics to simulate hidden pedestrians, enabling cautious but efficient behavior. End-to-end approaches like InterFuser~\cite{shao2023sinterfuser} have demonstrated strong performance and rank among the top performers on the CARLA Leaderboard.

Existing methods in this field rely on simplified simulation environments that fail to capture the complexity of real-world pedestrian behavior. HABIT addresses this gap by enabling realistic simulations in CARLA through semantically filtered motion-capture data, grounded trajectory reconstruction, and structured scenario generation.

%%%%%%%%%%%%%%%%%%%%%%%%%%%%%%%%%%%%%%%%%%%%%%%%%%%%%%%%%%%%%%%%%%%%%%%%%%%%%%%%%%%%%%%%%%%%%%%%%%%%%%%%%%%%%%%%%%%%%%%%%%%

\section{Methodology}
Figure~\ref{fig:architecture} summarizes our pipeline. Motion data is processed through one of two independent pathways. The primary path uses the AMASS datasets~\cite{Mahmood19}, where sequences pass through the \textit{Raw pose processing} and \textit{Vector extraction} modules to obtain 3D joints, pose vectors, and velocities. These are integrated via the \textit{Trajectory Reconstruction Module} to reconstruct global root trajectories aligned with local motion.

Additionally, pedestrian motions are extracted from open-source videos. These are processed through a \textit{Video Encoder} to extract 2D keypoints, which are lifted to SMPL format. Regardless of the source, using \textit{SMPLify} \cite{Bogo:ECCV:2016} and a temporal variant for videos, the keypoints are lifted to an intermediate SMPL representation. The resulting SMPL sequences are sampled to match CARLA’s target animation frequency and passed to the \textit{Retargeting to CARLA} module for conversion into CARLA-compatible joint rotations. In the following sections, we explain our pipeline in detail.

\subsection{Dataset filtering and processing}
We begin by collecting raw human motion data from a variety of heterogeneous motion capture datasets from AMASS and open-source videos. These datasets capture a broad range of everyday activities, not originally tailored for traffic contexts, but offer diverse kinematics and pose information crucial for realistic pedestrian simulation. The initial motion pool contains approximately $30,000$ general human activities with associated metadata including natural language descriptions and temporal structure, sourced from the HumanML3D motion-language dataset~\cite{Guo_2022_CVPR}, and the HumanSensing Lab~\cite{Panev_2024_WACV} text-annotated dataset.

To adapt these general-purpose motions for autonomous driving scenarios, we apply a two-stage filtering pipeline designed to select behaviorally plausible pedestrian motions suitable for traffic environments. First, we process the textual motion annotations from HumanML3D using the spaCy \textit{en\_core\_web\_sm} model, an industrial-strength natural language processing library, performing keyword filtering for walking and movement-related terms. This step reduces the dataset from $30,000$ motions to $5,108$ candidate motions appropriate for traffic contexts. The resulting semantically filtered set then undergoes a structured visual validation process, serving both as a human verification step and as an additional filter to remove motions irrelevant to traffic scenarios. The final curated set consists of $4730$ filtered motions in our current installation of the HABIT benchmark.

The resulting subset of filtered motions is then subjected to a standardized pre-processing pipeline. We begin by extracting joint positions from the AMASS pose data using a modified version of the raw pose processing technique introduced in HumanML3D. In addition to standard parameters, we compute linear and angular velocities of the root joint in a 6D rotation representation. These features are down-sampled to ensure temporal consistency with all the motions at $20$Hz. Intermediate SMPL pose representations are then estimated using a pretrained SMPLify network. Finally, the computed velocities are leveraged to reconstruct global root trajectories and orientations (see Section~\cref{sub:reconstruction}), which are subsequently aligned with the coarse SMPL poses derived from the joint data to yield consistent full-body motion representations.

\subsubsection{Motion from arbitrary videos}

We designed a Video-to-CARLA pipeline inspired by recent SMPL lifting pipelines such as WHAM~\cite{shin2024wham}. Our approach uses the ViTPose to detect 2D keypoints. The 2D keypoints are then uplifted into 3D SMPL poses using SMPLify. The global trajectory reconstruction algorithm described in \cref{sub:reconstruction}, is complemented with DROID-SLAM\cite{teed2021droid} to use the estimated camera intrinsics to project trajectories into global reference frame. 

We process a diverse set of static-camera videos sourced from YouTube \cite{youtube_website}, Pexels \cite{pexels_website}, and the HumanSensing Lab dataset~\cite{Panev_2024_WACV}, which provides frame-wise textual annotations. While the HABIT toolkit includes this processing framework, our goal here is to demonstrate HABIT’s extensibility for creating novel benchmarks and cross-domain applications.

From the HumanSensing Lab dataset, we select a subset of $75$ short, stationary pedestrian motions, however they are not a part of HABIT benchmark as they are positioned outside interaction range. These motions depict ambient sidewalk traffic in small groups and do not involve vehicle–pedestrian interactions. Importantly, these stationary motions differ from HABIT's \textit{idle} motions: whose primary purpose is to augment interactive sequences by extending scenario duration and enriching context with naturalistic gestures, and not just sidewalk traffic.

\subsection{Trajectory Reconstruction}
\label{sub:reconstruction}

We reconstruct physically plausible global root trajectories from local motion representations of the SMPL joints. Each motion sequence consists of a per-frame 6D root orientation vector $\mathbf{r}_t \in \mathbb{R}^6$ and a root-relative translational velocity vector $\mathbf{v}_t \in \mathbb{R}^3$.

The 6D root orientation vectors are converted into valid rotation matrices $\mathbf{R}_t \in \text{SO}(3)$, the special orthogonal group of 3D rotations, through a continuous 6D-to-rotation matrix projection~\cite{zhou2019continuity}. This step ensures smooth, gimbal-free orientation over time and forms the basis for global motion integration.

% \begin{figure}[h]
%   \centering
%   \includegraphics[width=0.9\linewidth]{sec/figures/global_root.png}
%   \caption{Global root}
%   \label{fig:global_root}
% \end{figure}

\begin{figure}[h]
  \centering
  \includegraphics[width=1\linewidth]{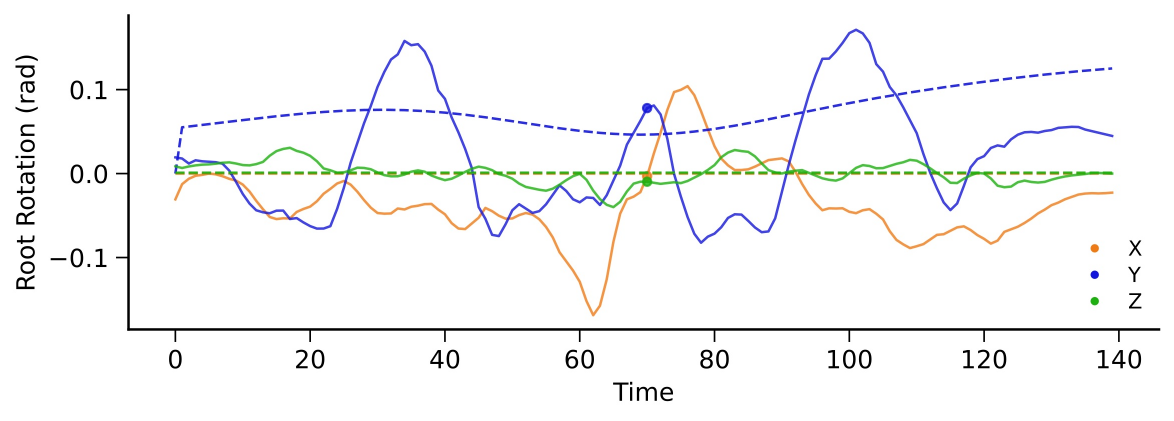}
  \caption{Global root rotation over time: original (solid) vs. reconstructed (dashed).}
  \label{fig:root-rot}
\end{figure}

To recover the global root trajectory, we rotate each local velocity vector into the world frame:
\begin{equation}
    \mathbf{v}^{\text{world}}_t = \mathbf{R}_t \cdot \mathbf{v}_t
\end{equation}
and compute the global translation at time $t$ via cumulative integration:
\begin{equation}
    \mathbf{T}_t = \mathbf{T}_0 + \sum_{\tau=1}^{t} \mathbf{v}^{\text{world}}_\tau
\end{equation}

where $\mathbf{T}_0 \in \mathbb{R}^3$ is the initial root position, typically set to the origin. 
This velocity-aligned trajectory reconstruction synchronizes global translation with local orientation at each timestep, enforcing temporal consistency by integrating motion along the pedestrian’s forward direction rather than using frame-wise deltas (see Figure~\ref{fig:root-rot}). It effectively reduces common SMPL artifacts such as abrupt heading changes, foot sliding, and inconsistent velocities, resulting in stable, realistic trajectories well-suited for simulation frameworks like CARLA.
%This velocity-aligned trajectory reconstruction aligns the pedestrian’s global translation with its local orientation at each timestep. This orientation-aware integration inherently enforces temporal consistency by accumulating displacement along the pedestrian's forward-facing direction, rather than relying on frame-wise displacement deltas. As a result, it mitigates artifacts common in SMPL-extracted trajectories, such as abrupt heading shifts, foot sliding, or inconsistent velocity magnitudes. The resulting global translations exhibit stable, physically plausible motion patterns that reflect natural pedestrian dynamics and support robust deployment in downstream simulation frameworks such as CARLA.

% \begin{figure}[h]
%   \centering
%   \includegraphics[width=0.9\linewidth]{sec/figures/traj.png}
%   \caption{Trajectory}
%   \label{fig:traj-good}
% \end{figure}

\subsection{SMPL-to-CARLA Motion Retargeting}
\label{retargeting-module}

\begin{figure}[h]
  \centering
  \begin{subfigure}[b]{0.48\linewidth}
  \centering
    \includegraphics[width=0.75\linewidth]{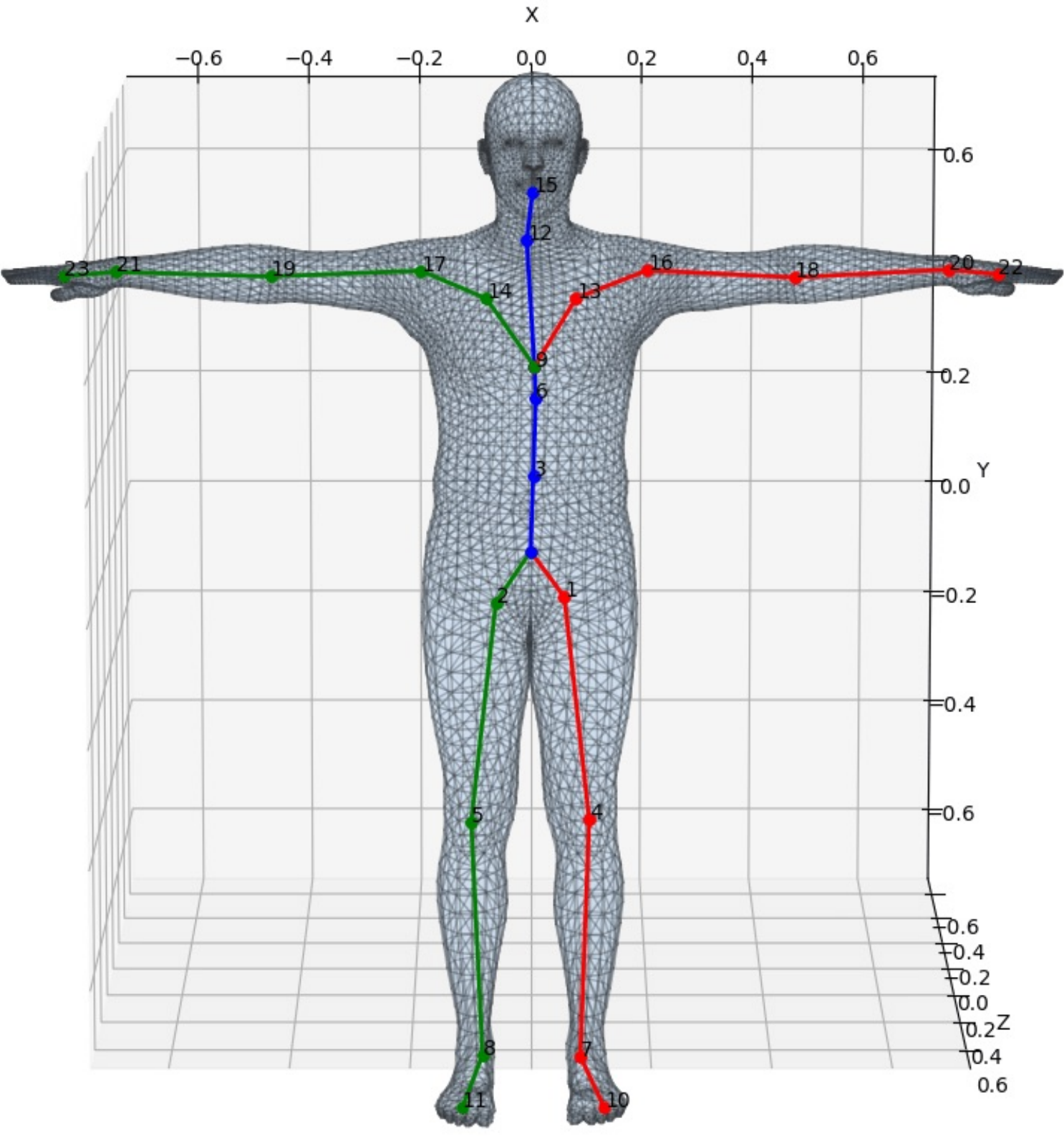}
    \subcaption{SMPL zero pose}
    \label{fig:smpl}
  \end{subfigure}
  \hfill
  \begin{subfigure}[b]{0.48\linewidth}
  \centering
    \includegraphics[width=0.75\linewidth]{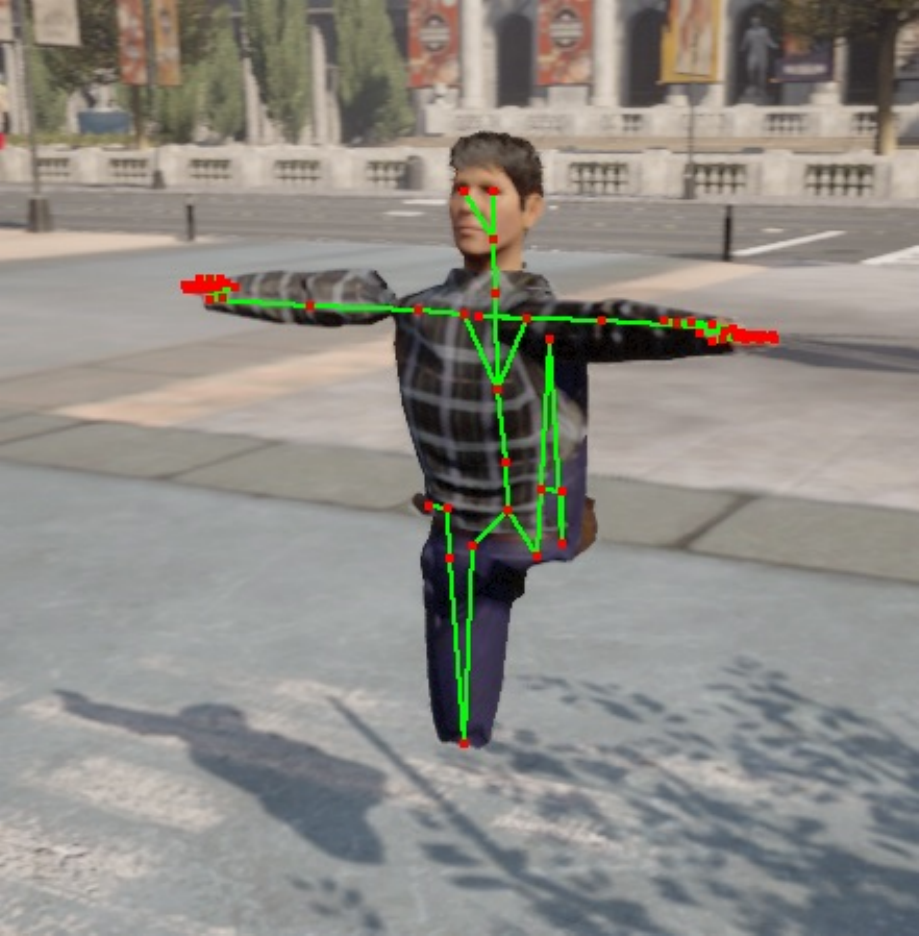}
    \subcaption{CARLA zero pose}
    \label{fig:carla_t}
  \end{subfigure}
  \caption{Canonical poses of SMPL~\cite{9417684} and CARLA skeletons~\cite{carla_issue7621}. Note the difference in joint definitions, and limb angles}
  \label{fig:pose_compare}
\end{figure}

SMPL provides human pose sequences as axis-angle representations $\boldsymbol{\theta} \in \mathbb{R}^{T \times J \times 3}$, defined per joint in a right-handed coordinate system. In contrast, CARLA, based on Unreal Engine~\cite{unrealengine}, employs a left-handed convention and supports only local joint rotations as Euler angles relative to a predefined skeletal rest pose. Notably, CARLA’s rest pose is neither a standard T-pose nor an A-pose, but rather a deformed body configuration, as illustrated in Figure~\ref{fig:pose_compare}. To reconcile these discrepancies, we design a motion retargeting pipeline that maps SMPL poses into CARLA-compatible Euler joint rotations $\boldsymbol{\phi} \in \mathbb{R}^{T \times J \times 3}$.

Each SMPL joint rotation vector $\boldsymbol{\omega}_{t,j} \in \mathbb{R}^3$ is first converted to a rotation matrix $R_{t,j} \in \mathrm{SO}(3)$ via the exponential map:
\begin{equation}
    R_{t,j} = \exp\left(\hat{\boldsymbol{\omega}}_{t,j}\right),
\end{equation}
where $\hat{\boldsymbol{\omega}}$ denotes the skew-symmetric matrix and the exponential map $\exp(\hat{\boldsymbol{\omega}})$ yields a valid 3D rotation.

To adapt SMPL's right-handed rotations into CARLA’s left-handed frame, we apply a global coordinate transformation $C \in \mathrm{SO}(3)$:
\begin{equation}
    R^{\text{CARLA}}_{t,j} = C R_{t,j} C^{-1},
\end{equation}
where $C$ is constructed to flip the forward and up axes to align with CARLA's coordinate basis.

As CARLA defines a custom skeleton, we map joints based on most suitable closest match and certain SMPL joints must be merged to cater to the simplified representation in CARLA. For example, CARLA’s \texttt{Spine1} is computed from the chained rotations of SMPL’s \texttt{Spine2} and \texttt{Spine3}:
\begin{equation}
    R^{\text{CARLA}}_{t,\texttt{Spine1}} = R^{\text{CARLA}}_{t,\texttt{Spine2}} \cdot R^{\text{CARLA}}_{t,\texttt{Spine3}}.
\end{equation}

CARLA expects local joint rotations expressed relative to deformed zero pose, which we offset using a pre-defined standard T pose offsets for the CARLA skeleton, creating a new reference T pose for CARLA skeleton, $R^{\text{ref}}_j \in \mathrm{SO}(3)$. We compute the motion-relative joint rotation $\Delta R_{t,j}$ via:
\begin{equation}
    \Delta R_{t,j} = \left(R^{\text{ref}}_j\right)^{-1} R^{\text{CARLA}}_{t,j} R^{\text{ref}}_j,
\end{equation}
centering rotations around the reference pose. The final Euler joint angles are extracted using the intrinsic XYZ convention:
\begin{equation}
    \boldsymbol{\phi}_{t,j} = \operatorname{Euler}_{\text{XYZ}}(\Delta R_{t,j}),
\end{equation}
yielding $\boldsymbol{\phi} \in \mathbb{R}^{T \times J \times 3}$ suitable for direct animation in CARLA.

%Finally, to support evaluation of autonomous agents under realistic human motion, we integrate our retargeted pedestrian motions into CARLA’s \texttt{Leaderboard} framework and simulate traffic interactions.

%%%%%%

\section{HABIT Benchmark}
\label{sec:bench}
The curated set of $4,730$ motion sequences obtained by our two-stage filtering pipeline is additionally annotated for structured scenario generation and benchmarking, retaining original natural language descriptions and unique identifiers. High-level behavior categories (e.g., walking, running, standing, falling) are inferred via NLP clustering, enabling reproducible behavior conditioning and semantic role assignment in simulation.

The current release of the HABIT benchmark comprises $110$ routes under $12$ distinct weather conditions, with each scenario populated by $30$ vehicles, $20$ behaviorally diverse pedestrians, and $10$ ambient pedestrians serving as meaningful and background traffic. 

%The curated set of 4,730 motion sequences, produced through the filtering pipeline, is annotated to support the generation and benchmarking of structured scenarios. See Figure~\ref{fig:method_placeholder}. Each motion retains its original natural language description and is assigned a unique identifier. A high-level behavioral category is automatically inferred for each sequence by natural language processing to map the textual descriptions to pedestrian behavior classes, such as walking, running, standing, or falling. These annotations enable reproducible behavior conditioning and facilitate automatic assignment of semantic roles in simulation. 
%With the processed CARLA motion data, accompanied by the structural annotations and behavioral tags, the simulation benchmark procedures are fully prepared for implementation.
\begin{figure}[h]
  \centering
  \includegraphics[width=\linewidth, keepaspectratio]{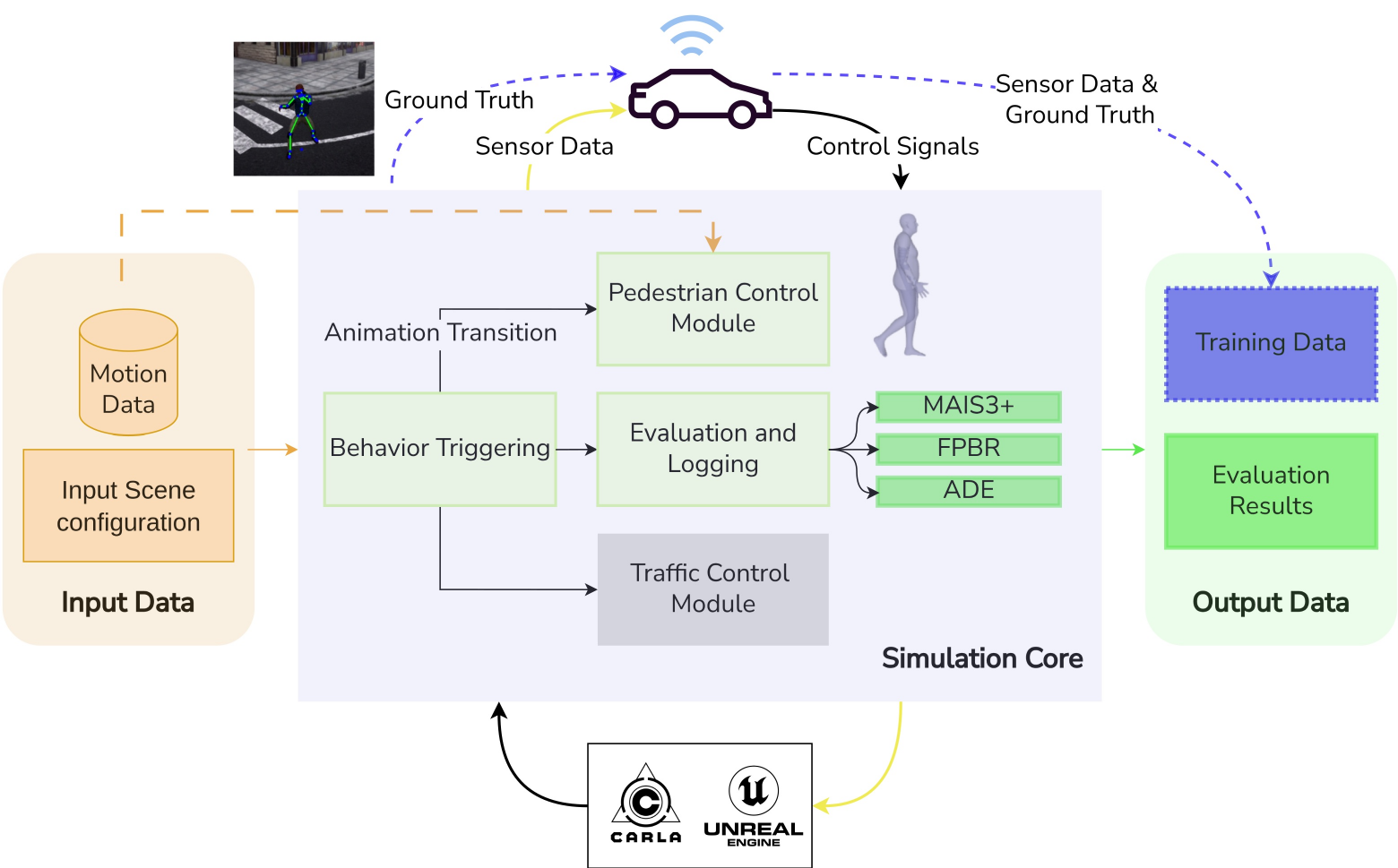}
  \caption{Overview of the HABIT benchmark pipeline where retargeted motions are placed and controlled into CARLA scenes.}
  \label{fig:method_placeholder}
\end{figure}

\subsection{Simulation Scenario Generation}

To evaluate autonomous agents under realistic human motion, we integrate the curated pedestrian dataset into interactive scenarios within the CARLA simulator (Figure~\ref{fig:method_placeholder}). Each scenario features an ego vehicle navigating urban routes and encountering pedestrians exhibiting one of three behavior categories: not crossing, attempting to cross, or crossing. These categories are determined by forward displacement thresholds to preserve behavioral intent. The scenario generation pipeline defines agent placement, motion assignment, and interaction triggers for structured yet diverse evaluations within the CARLA Leaderboard. 

\subsubsection{Agent Configuration and Interaction Logic}

Each simulation scenario consists of a single ego vehicle following a predefined route, controlled by CARLA’s autopilot or external planners, and a fixed number of pedestrian agents. Background traffic behavior is governed by CARLA's ScenarioRunner module.

Pedestrians are pre-spawned at semantically appropriate locations such as sidewalks and curbs. Spawn points are randomly selected from CARLA’s static sidewalk meshes within a bounded region around the ego route, ensuring pedestrian-viable placements through semantic tagging.

Initially, pedestrian agents exhibit idle, low-displacement motions (e.g., gesturing, scanning, subtle posture shifts), enhancing realism while ensuring scene stability. Behavior transitions to dynamic goal-oriented actions (e.g., crossing, attempting to cross) occur based on the ego vehicle’s proximity and orientation or scenario-level triggers compatible with CARLA Leaderboard specifications. To maintain visual continuity, transitions employ short blending sequences interpolating between the initial idle pose and the dynamic motion start pose.

Pedestrian trajectories are continuously monitored for potential collisions with static objects (e.g., walls, parked vehicles) using bounding-box intersection checks. Predicted collisions prompt real-time trajectory rerouting to preserve locomotion and intent. Dynamic collisions with moving vehicles are permitted and logged to assess safety-critical scenarios. This selective collision logic maintains environmental realism and consistency for planner evaluation.

\subsection{Evaluation Metrics}

Our benchmark integrates all official evaluation metrics from the CARLA Leaderboard, along with two additional metrics designed to assess the behavioral and safety implications of pedestrian-agent interactions.

\emph{Injury Severity (AIS 3+):} To provide a more informative assessment than binary collision counts, we incorporate a clinically grounded injury severity metric based on the Abbreviated Injury Scale (AIS)~\cite{hu2023simulation}. Specifically, we estimate the probability of a serious injury (MAIS 3+, i.e., AIS level 3 or higher) using a logistic regression model from Yanagisawa \etal~\cite{yanagisawa2017estimation}, which relates injury risk to pedestrian impact speed $v$ (in m/s):
\begin{equation}
P_{\text{MAIS 3+}}(v) = \frac{1}{1 + \exp(3.164 - 0.288 v)}
\end{equation}

\emph{False Positive Braking Rate (FPBR):} We propose FPBR as a metric to quantify overcautious behavior in planners that incorporate pedestrian trajectory forecasting. FPBR is defined as the ratio of braking events that occur without an actual pedestrian crossing to the total number of braking events. This metric specifically targets systems like InterFuser~\cite{shao2023sinterfuser} and PlanT~\cite{Renz2022CoRL}, where the planner uses predicted pedestrian trajectories to assess collision risk and trigger braking. FPBR highlights unnecessary slowdowns caused by inaccurate or overly conservative forecasts.

Together, these metrics offer a multi-faceted evaluation, capturing both classical performance metrics and finer-grained behavioral sensitivity to complex pedestrian motion.

\subsection{Dataset Statistics}

\begin{figure}[h]
  \centering
  \includegraphics[width=\linewidth, keepaspectratio]{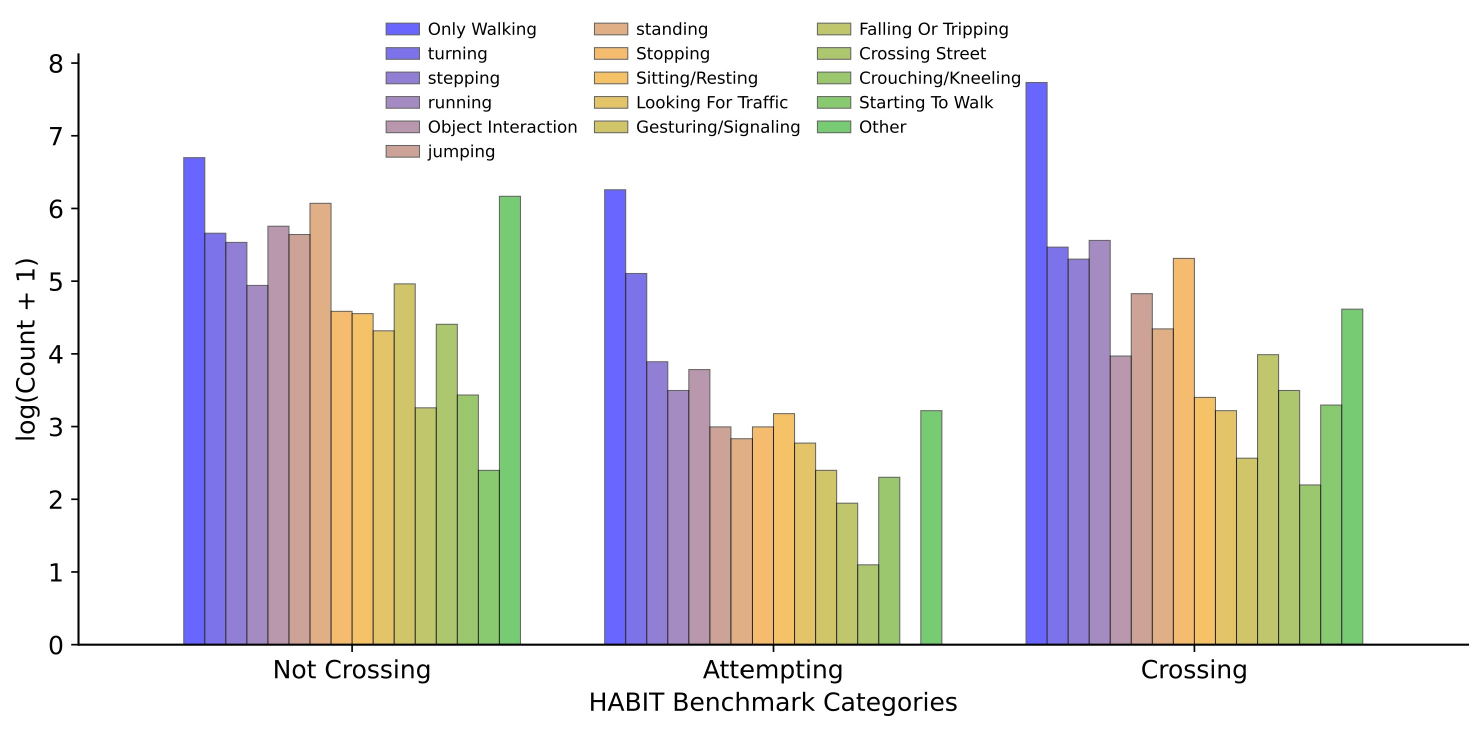}
  \caption{Distribution of motion tags by behavior category.}
  \label{fig:stats}
\end{figure}

The final benchmark includes 4,730 curated motion sequences, representing a diverse and balanced range of core pedestrian behaviors. To ensure comprehensive coverage, these behaviors are grouped into three broad classes, with NLP-derived behavioral tags proportionally distributed across them. Figure~\ref{fig:stats} visualizes this distribution, providing a clear overview of the dataset’s behavioral diversity.

\begin{figure}[h]
  \centering
  \includegraphics[width=\linewidth, keepaspectratio]{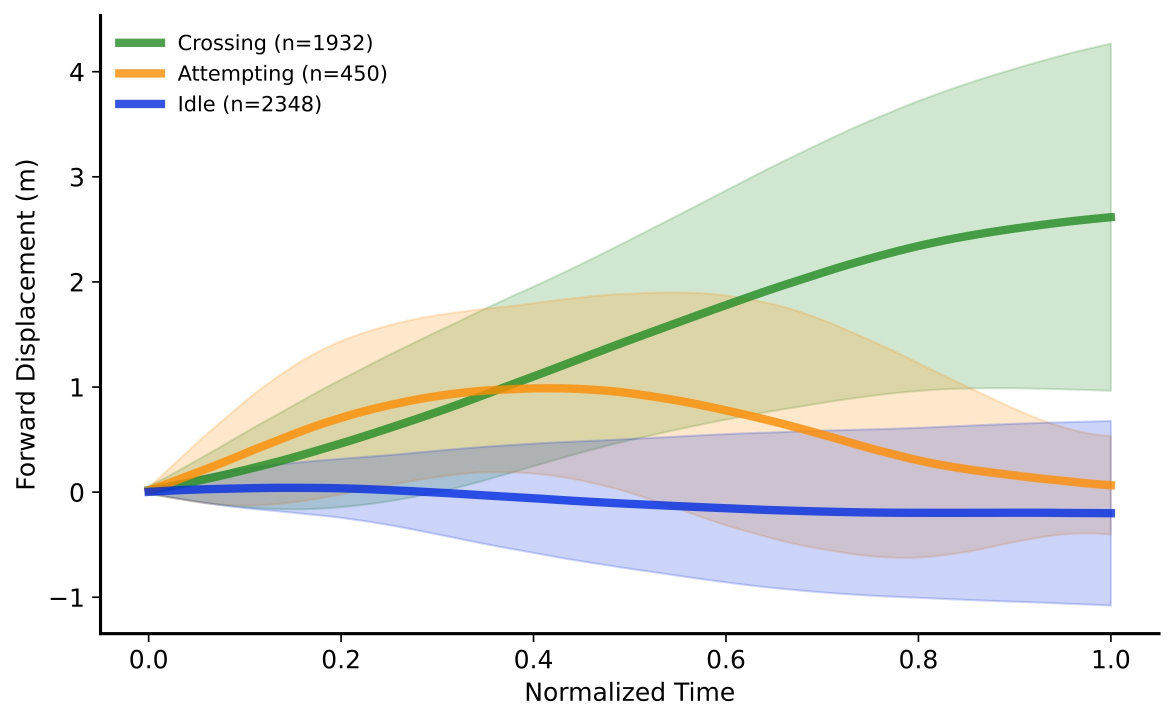}
  \caption{Trajectory statistics (mean, variance) by behavior class: Not Crossing, Attempting, Crossing.}
  \label{fig:mean-disp}
\end{figure}

To characterize the dataset, we compute mean representative motions for each of the three main behavior categories. These averaged trajectories capture prototypical pedestrian dynamics and are visualized in Figure~\ref{fig:mean-disp}, highlighting distinct patterns and variability, especially among crossing behaviors.

\section{Experiments}
\label{sec:experiments}

To demonstrate the realism and utility of the HABIT benchmark, we conduct comprehensive evaluations on driving models. Our goal is to analyze the impact of diverse, semantically grounded pedestrian behavior on planning robustness and prediction accuracy of autonomous agents. Prior to HABIT, such evaluations were either infeasible or uninformative on synthetic pedestrian data due to its limited behavioral diversity and visual domain gap. By scaling up realism in the CARLA simulator, HABIT bridges this gap, enabling meaningful benchmarking of pedestrian-aware autonomy under conditions that better reflect the complexities of the real world.

\subsection{Prediction and Planning on HABIT}
\label{sec:predFull}
% We evaluate the robustness of \textit{InterFuser}~\cite{shao2023sinterfuser}, a transformer-based autonomous driving agent integrating perception, prediction, and planning, on HABIT's semantically diverse pedestrian behaviors. Although InterFuser achieves zero collisions/km on the CARLA Leaderboard, featuring predictable pedestrian scripts, its performance significantly degrades under HABIT’s richer and more ambiguous pedestrian motions.
We evaluate three recent end-to-end autonomous driving agents on HABIT's semantically diverse pedestrian behaviors. Specifically, we include \textit{InterFuser}~\cite{shao2023sinterfuser}, a transformer-based model that 
integrates perception, prediction, and planning; \textit{TransFuser}~\cite{chitta2022transfuser}, a transformer-based fusion architecture that jointly reasons over camera and LiDAR inputs; and \textit{BEVDriver}~\cite{winter2025bevdriver}, which leverages latent bird’s-eye-view representations together with large language models (LLMs) for trajectory planning. For the latter, we adopt the route-to-language module introduced in the LangAuto benchmark~\cite{shao2024lmdrive}, which encodes navigation waypoints into natural language form to condition planning. This evaluation enables a systematic comparison of how these agents generalize to richer and more ambiguous pedestrian motions than those found in scripted benchmark scenarios.

\begin{table}[h]
  \centering
  \resizebox{\linewidth}{!}{
    \begin{tabular}{@{}lcccc@{}}
      \toprule
      \textbf{Model} & \textbf{Collisions/km} $\downarrow$ & \textbf{pMAIS3+ (\%)} $\downarrow$ & \textbf{FPBR} $\downarrow$ & \textbf{ADE} $\downarrow$ \\
      \midrule
        InterFuser & 5.24 & 10.96 & 0.33 & 1.64 \\
        TransFuser & 7.43 & 12.94 & 0.12 & 1.87 \\
        BEVDriver  & 7.19 & 12.35 & - & - \\
      \bottomrule
    \end{tabular}
  }
  \caption{Performance of three end-to-end planners on \textbf{HABIT Benchmark}. Metrics: safety (collisions/km, pMAIS3+), conservatism (FPBR), and pedestrian prediction (ADE). Lower is better; higher pMAIS3+ = more severe injuries.}
  \label{tab:habit_full_interfuser_bevdriver}
\end{table}

Results in Table~\ref{tab:habit_full_interfuser_bevdriver} highlight that HABIT reveals planner-specific weaknesses that remain invisible in scripted benchmarks. Most notably, InterFuser, despite achieving zero collisions/km on the CARLA Leaderboard, records \textbf{5.24 collisions/km} and a \textbf{10.96\%} injury risk under HABIT. This sharp degradation underscores a critical limitation: its planner and controller, trained on rule-based pedestrians, adopt an overly conservative braking strategy (high FPBR of 0.33) yet still fails to anticipate diverse pedestrian motions, resulting in frequent collisions. In contrast, TransFuser displays the opposite tendency, with fewer unnecessary braking events (low FPBR of 0.12) but correspondingly higher collision frequency and prediction error, indicating underreactive planning. BEVDriver achieves safety scores comparable to TransFuser but lacks pedestrian intent forecasting, making FPBR and ADE evaluation inapplicable.

Taken together, these findings demonstrate that HABIT systematically stresses planning agents with realistic pedestrian behavior, surfacing trade-offs between safety and conservatism. In particular, the large performance drop of InterFuser emphasizes the need for benchmarks like HABIT to expose hidden biases and generalization failures in state-of-the-art driving agents.

\subsection{Ablation studies}

To isolate the effects of specific pedestrian behaviors, we evaluate InterFuser on two HABIT variants: (i) \textbf{HABIT-Idle}, where all pedestrians remain idle with no crossing intent, and (ii) \textbf{HABIT-Left}, where pedestrians exclusively originate from the left side and attempt rightward crossings.

\begin{figure}[h]
  \centering
  \includegraphics[width=1\linewidth]{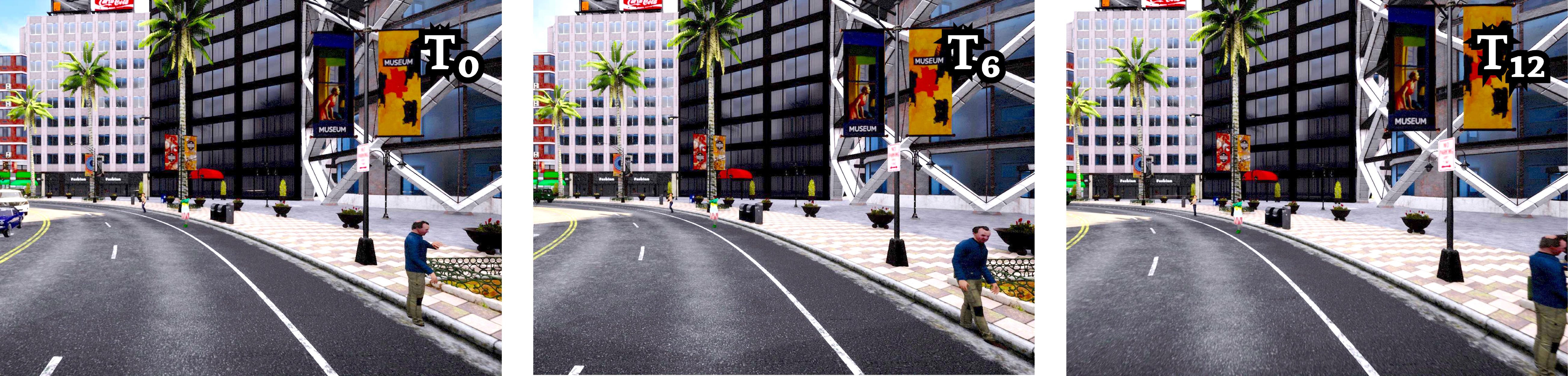}
  % \caption{\textbf{False Positive Braking over Time.} Front-view camera frames at three timesteps ($T_0$, $T_6$, $T_{12}$) illustrate the planner’s overcautious behavior. At $T_0$, the vehicle initiates a full stop in response to a nearby idle pedestrian. At $T_6$, midway through the stop, the pedestrian remains stationary with no crossing intent. At $T_{12}$, after a total halt of 12 seconds, the planner resumes motion}
  \caption{\textbf{False Positive Braking over Time.} Front-view frames ($T_0$, $T_6$, $T_{12}$, , in seconds) show overcautious stopping: vehicle halts for a stationary pedestrian and resumes after 12 s.}
  \label{fig:fpbr}
\end{figure}

%In Table~\ref{tab:interfuser_ablation_idle_left} we report performance of Interfuser across four key metrics: Collisions/km (collision frequency), pMAIS3+ (severe injury probability), FPBR (fraction of unnecessary braking), and ADE (Average displacement error for pedestrian trajectory prediction). Results highlight significant performance deterioration on the HABIT-Full benchmark, demonstrating vulnerabilities not captured by rule-based simulations. Conversely, benchmarking with HABIT-Idle shows increased conservative braking, emphasizing the necessity of behaviorally diverse evaluations to expose latent weaknesses in autonomous driving systems.
% Table 2: InterFuser ablation on HABIT-Idle vs. HABIT-Left
\begin{table}[h]
  \centering
  \resizebox{\linewidth}{!}{
    \begin{tabular}{@{}lcccc@{}}
      \toprule
      \textbf{Condition} & \textbf{Collisions/km} $\downarrow$ & \textbf{pMAIS3+ (\%)} $\downarrow$ & \textbf{FPBR} $\downarrow$ & \textbf{ADE} $\downarrow$ \\
      \midrule
      HABIT-Idle & 0.16 & 8.40  & 0.639 & 1.84 \\
      HABIT-Left & 8.35 & 12.19 & 0.09  & 2.19 \\
      \bottomrule
    \end{tabular}
  }
  \caption{InterFuser ablation: On \textbf{HABIT-Idle}, overly conservative behavior raises FPBR; on \textbf{HABIT-Left}, performance drops due to right-to-left bias from data imbalance.}
  \label{tab:interfuser_ablation_idle_left}
\end{table}
Results in Table~\ref{tab:interfuser_ablation_idle_left} show two complementary failure modes. Under HABIT-Idle, InterFuser records almost no collisions but a sharp increase in FPBR (0.639). This reflects an overcautious policy: the agent frequently brakes in response to idle pedestrians, misinterpreting their presence as crossing intent. In effect, the model treats standing humans as latent threats, revealing its reliance on detection cues without robust intent reasoning. Figure~\ref{fig:fpbr} illustrates this tendency, where the vehicle remains stopped for 12 seconds despite the pedestrian showing no crossing intent.

\begin{figure}[h]
  \centering
  \includegraphics[width=1\linewidth]{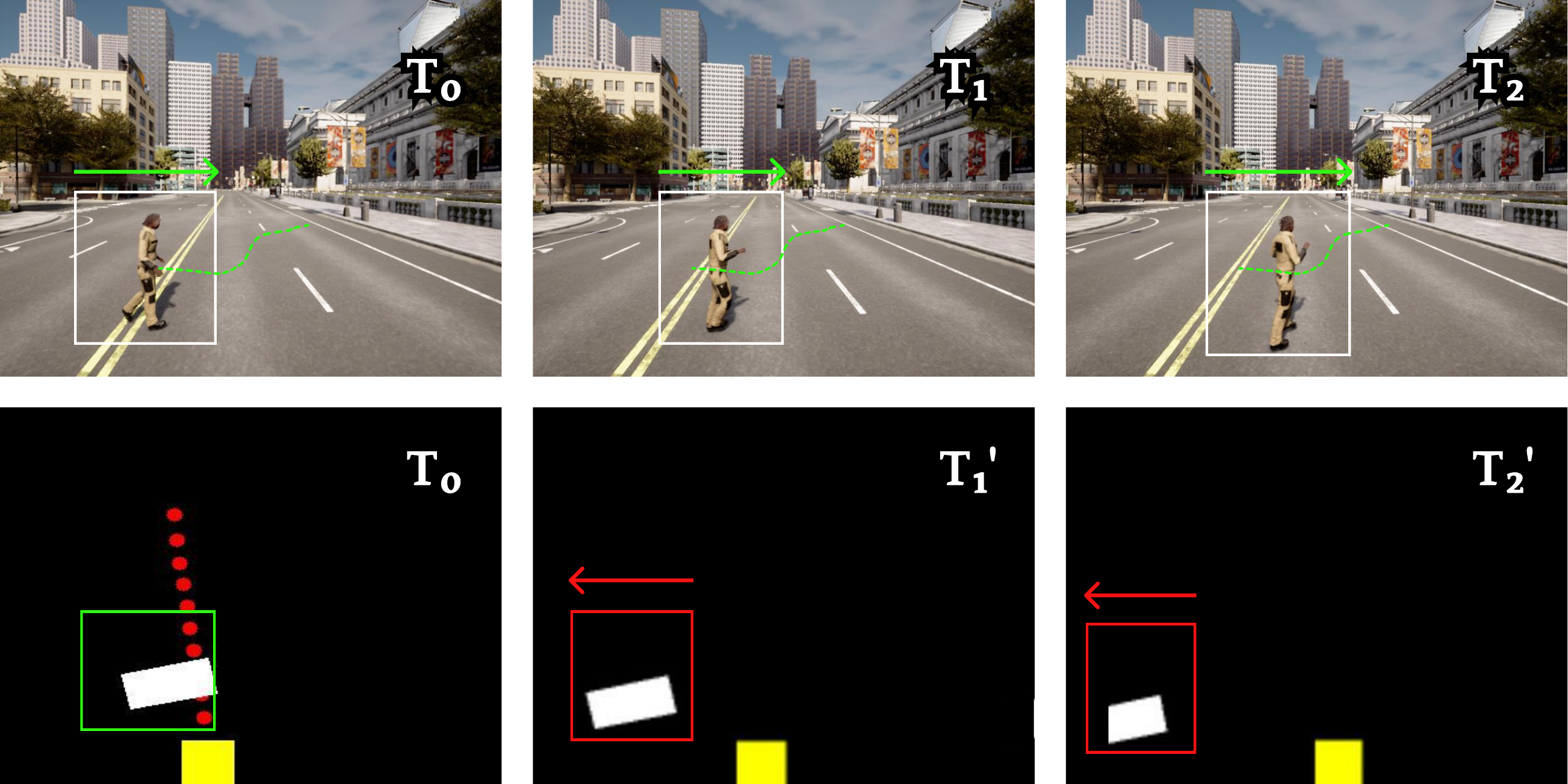}
  % \caption{\textbf{Drectional prediction bias (qualitative).} Top row: front camera input with ground truth trajectory of pedestrian (green) moving towards right side (for time step $T_0$--$T_2$). Bottom row: Predicted BEV maps at increasing timesteps. The model wrongly predicts the pedestrian going left, revealing directional bias}
  \caption{\textbf{Directional prediction bias (qualitative).} Top: front camera input with ground truth pedestrian trajectory (green) moving right ($T_0$--$T_2$, in seconds). Bottom: predicted BEV maps over time; model incorrectly predicts left, showing directional bias.}
  \label{fig:prediction_bias}
\end{figure}

In contrast, HABIT-Left exposes a strong directional bias. As illustrated in Figure~\ref{fig:prediction_bias}, InterFuser consistently forecasts right-to-left motion even when pedestrians cross from left to right. This misprediction, caused by data imbalance in its training distribution, leads to underreaction and late braking. Quantitatively, this results in \textbf{8.35 collisions/km} and a pMAIS3+ rate of \textbf{12.19\%}, alongside degraded prediction accuracy (ADE = 2.19). The low FPBR in this case indicates that the agent rarely brakes preemptively, further compounding safety risks.

% \begin{table}[h]
%   \centering
%   \resizebox{\linewidth}{!}{
%     \begin{tabular}{@{}lcccc@{}}
%       \toprule
%       \textbf{Condition} & \textbf{Collisions/km} $\downarrow$ & \textbf{pMAIS3+} $\downarrow$ & \textbf{FPBR} $\downarrow$ & \textbf{ADE} $\downarrow$ \\
%       \midrule
%       CARLA Leaderboard & \textcolor{red}{0.0} & - & - & - \\
%       HABIT (full)      & 5.24          & 10.96     & 0.33    & 1.64 \\
%       HABIT-Idle        & 0.16          & 8.40      & 0.639   & 1.84 \\
%       \bottomrule
%     \end{tabular}
%   }
%   \caption{InterFuser performance across three pedestrian configurations. Higher pMAIS3+ indicates more severe injuries. Lower values are preferred for all other metrics.}
%   \label{tab:habit_evaluation}
% \end{table}

Together, these ablations clarify why the performance degrades on the full HABIT benchmark (see Table~\ref{tab:habit_full_interfuser_bevdriver}): InterFuser oscillates between excessive caution in idle scenarios and unsafe underreaction when confronted with rare but realistic crossing behaviors. Our ablation results highlight the need for behaviorally diverse benchmarks to uncover planner biases that remain invisible in rule-based environments.

%Figure~\ref{fig:prediction_bias} illustrates a critical failure in InterFuser’s pedestrian forecasting. The model consistently predicts right-to-left trajectories, regardless of pedestrians' true intent, particularly evident when pedestrians cross from the left sidewalk. This directional bias, likely arising from data imbalance during training, leads to inadequate planning responses and late braking decisions.

%Evaluations on HABIT routes using exclusively left-origin pedestrian crossings reveal substantial performance degradation, resulting in 8.35 collisions/km and a pMAIS3+ injury severity rate of $12.19\%$, despite maintaining a low False Positive Braking Rate (FPBR) of 0.09. An Average Displacement Error (ADE) of 2.19 further highlights degraded prediction accuracy under spatially asymmetric pedestrian behaviors.

\section{Discussion}
\label{sec:discussion}

Our experiments show that HABIT substantially advances the realism and diagnostic depth of pedestrian-aware benchmarks for autonomous driving. For prediction and planning, HABIT revealed critical shortcomings in state-of-the-art models. Interfuser, despite strong performance on the CARLA Leaderboard, struggled with HABIT’s semantically rich scenarios, especially in anticipating crossing intents from uncommon spatial contexts. These failures, reflected in elevated collision rates, displacement errors, and MAIS 3+ injury risks, underscore the limitations of training on rule-based or scripted pedestrian behaviors. TransFuser faced similar difficulties, reacting less conservatively but with higher collision frequency and less accurate trajectory forecasts. BEVDriver performed comparably in safety terms but lacked explicit pedestrian forecasting, limiting its ability to capture fine-grained behavioral intent.

HABIT also enables fine-grained analysis of model weaknesses through controlled ablations. In idle-pedestrian scenarios, agents exhibited overcautious braking, while left-origin crossings exposed strong directional biases and systematic mispredictions. Together, these analyses show that HABIT surfaces subtle but safety-critical planner failures, highlighting trade-offs between conservatism and responsiveness. Its structured behavioral diversity ensures that planners are tested against realistic, unpredictable pedestrian interactions, rather than oversimplified patterns.
%HABIT also enables fine-grained analysis of model weaknesses through metrics like false-positive braking rate, injury severity, and collision frequency, effectively surfacing subtle but safety-critical planner failures. Its structured behavioral diversity ensures that planners are tested against realistic, unpredictable pedestrian interactions, rather than oversimplified patterns.
 %Zero-shot evaluations on perception tasks demonstrate HABIT’s high visual and biomechanical fidelity, with YOLOv11l-pose achieving competitive, real-world–comparable performance without simulation-specific fine-tuning. Similarly, the Segment Anything Model (SAM) produced stable segmentation masks despite being trained exclusively on real data, validating HABIT’s spatial coherence and motion continuity.

Despite this progress, challenges remain. Grounded motion extraction from arbitrary, moving-camera video sources is still difficult, requiring improved reconstruction techniques to maintain spatial accuracy. Additionally, traditional augmentation introduces redundant behaviors; future work could employ generative models to create richer intra-trajectory and pose variations~\cite{10588860}. Lastly, pedestrian spawning in simulation can be made more realistic by learning placement strategies from real-world data, enabling context-aware agent initialization. Addressing these areas will further enhance HABIT’s scalability and realism for autonomous system evaluation.

\section{Conclusion and future work}
In this work, we introduced HABIT (Human Action Benchmark for Interactive Traffic), a high-fidelity simulation benchmark designed to address the critical limitations of current autonomous driving (AD) simulations in representing realistic human behavior. HABIT integrates real-world pedestrian motions into CARLA, offering a rigorous and realistic platform for evaluating pedestrian-aware autonomous driving systems. Our evaluations revealed that while modern end-to-end planning methods can achieve zero collisions per kilometer on the CARLA Leaderboard, they incur up to $7.43$ collisions/km and a $12.94$\% AIS 3+ injury risk on HABIT, and unnecessarily brake in $33$\% of the cases. These findings highlight HABIT's ability to uncover critical failure modes and blind spots in pedestrian-aware autonomy that were missed by prior evaluations. Beyond AD, HABIT's pipeline offers a versatile solution for realistic human motion simulation, supporting embodied learning and enhancing human-agent interaction in robotics. To foster open research and reproducibility, the HABIT benchmark and its entire simulation pipeline are publicly released.

%We introduced HABIT, a high-fidelity simulation benchmark that integrates curated real-world human motion into CARLA for evaluating pedestrian-aware autonomous driving systems. In summary, HABIT offers a rigorous, realistic evaluation platform that not only bridges the simulation-to-reality gap but also uncovers critical blind spots in pedestrian-aware autonomy. Beyond autonomous driving, the HABIT pipeline is a versatile solution for realistic human motion simulation. It can support embodied learning, enhance human-agent interaction in robotics, and improve training data for generative human motion models, offering a rigorous platform for safety-critical, interactive, and data-driven applications. To support open research and reproducibility, the HABIT benchmark will be publicly released in accordance with CARLA's dataset guidelines, and the entire simulation pipeline will be open-sourced for the community. We are committed to continually improving the framework—by developing more accurate motion retargeting methods and incorporating generative motion models—toward building a fully customizable and extensible benchmarking tool for pedestrian simulation at scale.

{
    \small
    \bibliographystyle{unsrtnat}
    \bibliography{main}
}
\clearpage
\appendix
\section*{\Large Appendix}

%%%%%%%%% BODY TEXT - ENTER YOUR RESPONSE BELOW
\section{HABIT Benchmark Illustrations and Details}
\begin{figure}[htbp]
  \centering
  \includegraphics[width=\linewidth, height=0.9\linewidth]{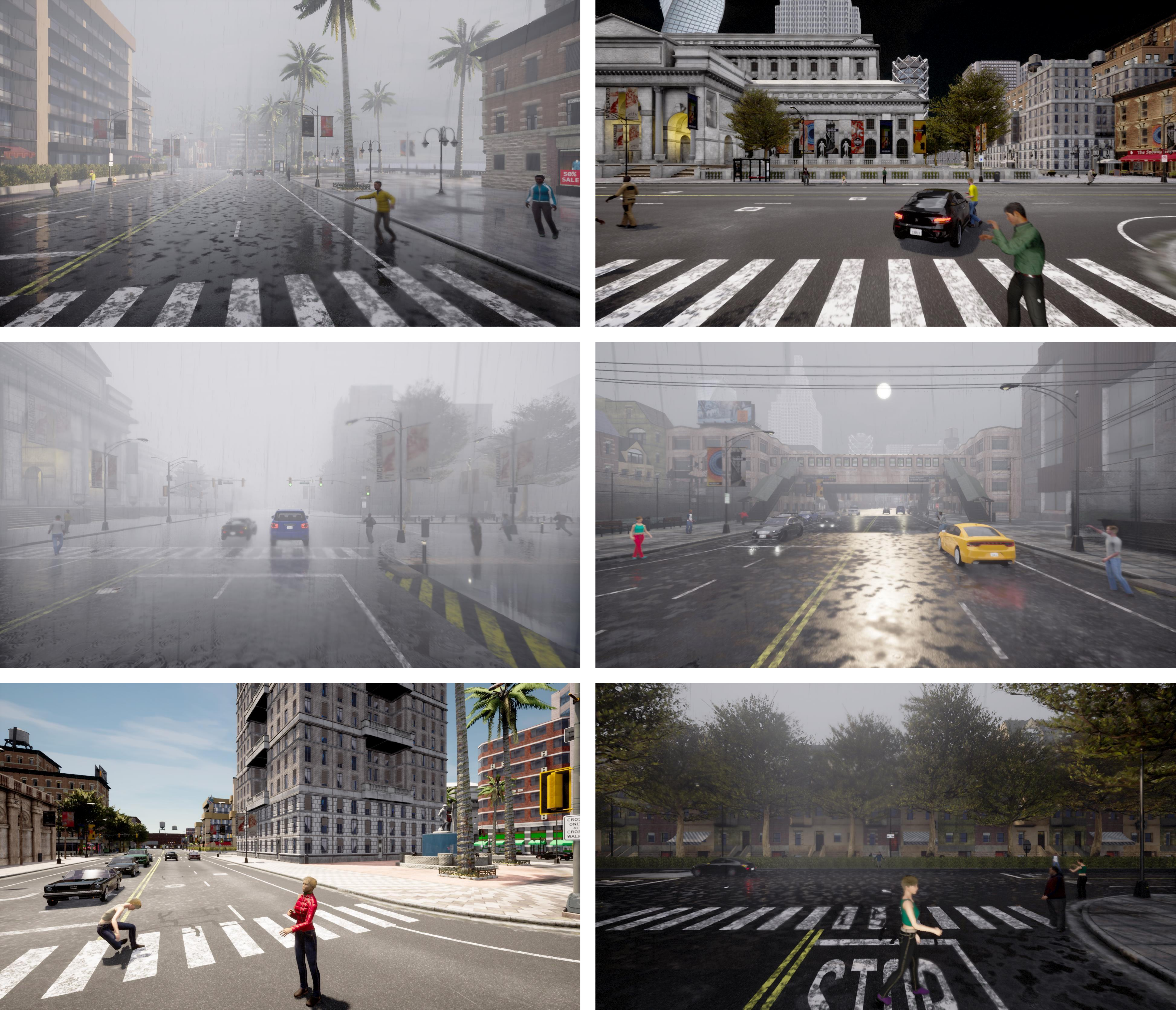}
  \caption{
    \textbf{HABIT Benchmark.} 
    Examples illustrating diverse pedestrian behaviors, environmental conditions, and route configurations designed to bridge the reality gap. 
  }
  \label{fig:habit_benchmark}
\end{figure}

The primary objective of the HABIT benchmark is to assess the driving competence of autonomous agents in complex and realistic traffic environments, with particular emphasis on interactions involving vulnerable road users such as pedestrians. A key limitation of existing benchmarks is their insufficient representation of pedestrians performing rare or unconventional gestures and behaviors. To address this gap, our initial evaluations focus on state-of-the-art agents from the CARLA leaderboard ~\cite{Leaderboard24}. The current release of the HABIT benchmark comprises $110$ routes under $12$ distinct weather conditions, with each scenario populated by $30$ vehicles, $20$ behaviorally diverse pedestrians, and $10$ ambient pedestrians serving as meaningful and background traffic. Representative examples from the benchmark are presented in Figure~\ref{fig:habit_benchmark}.

\section{Comparison of HABIT with AD Simulators}

\begin{figure}[!h]
    \centering
    \begin{minipage}[t]{0.32\linewidth}
        \centering
        \includegraphics[width=\linewidth]{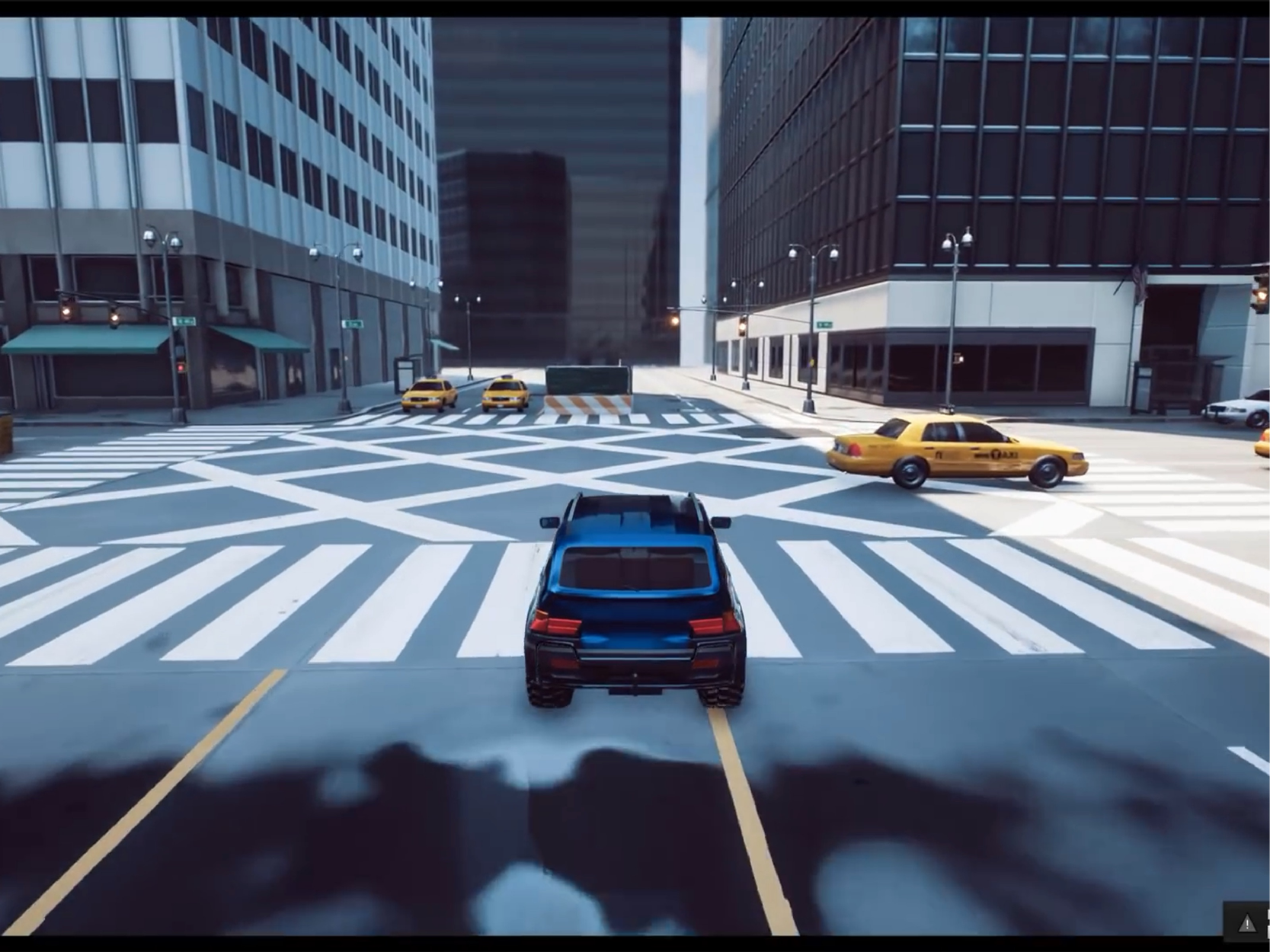} \\
        \textbf{Airsim}\\
        \raggedright \footnotesize No native pedestrian asset class
    \end{minipage}%
    \hfill
    \begin{minipage}[t]{0.32\linewidth}
        \centering
        \includegraphics[width=\linewidth]{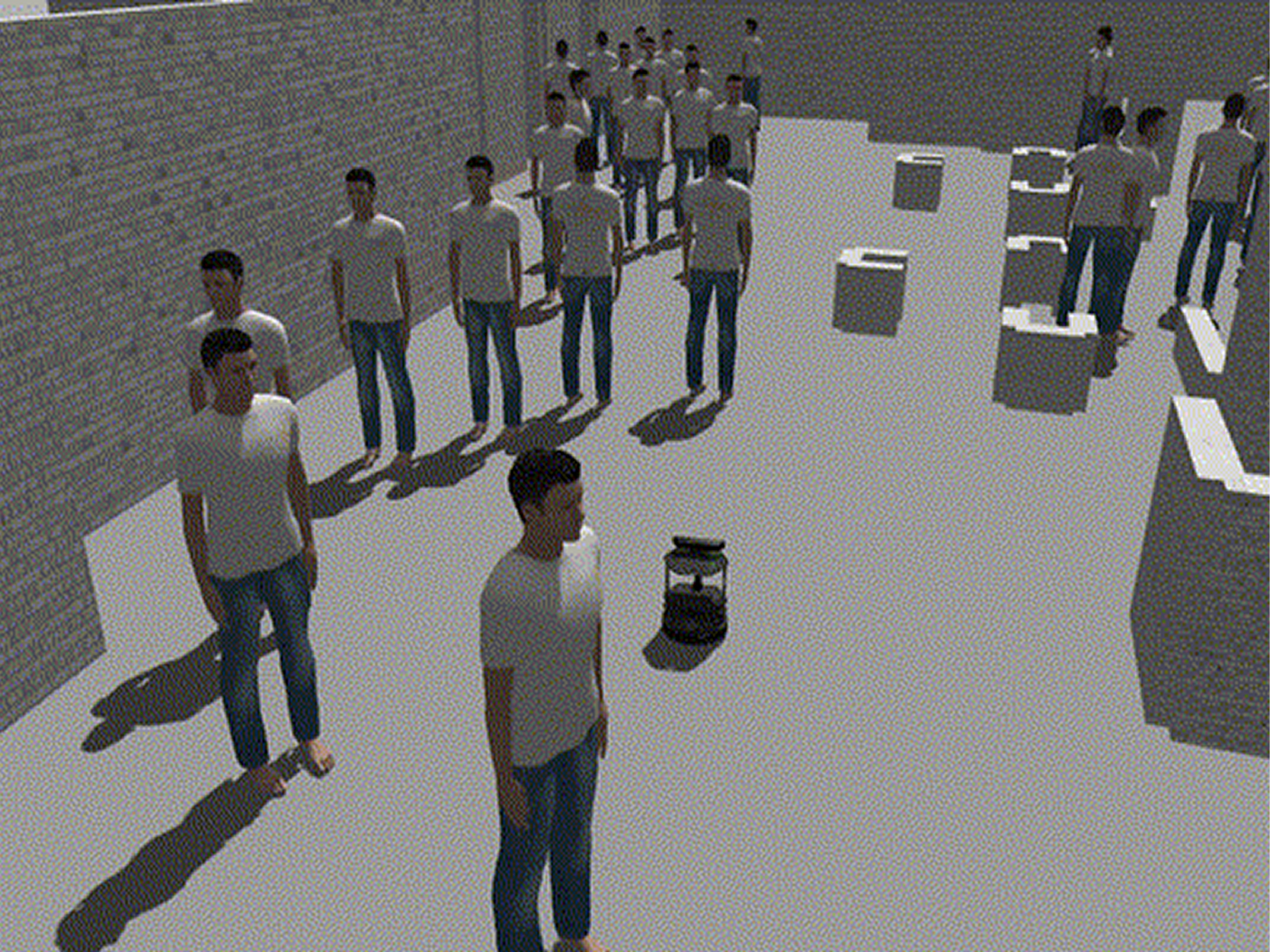} \\
        \textbf{Gazebo}\\
        \raggedright \footnotesize Social-force pedestrian model, no gestures
    \end{minipage}%
    \hfill
    \begin{minipage}[t]{0.32\linewidth}
        \centering
        \includegraphics[width=\linewidth]{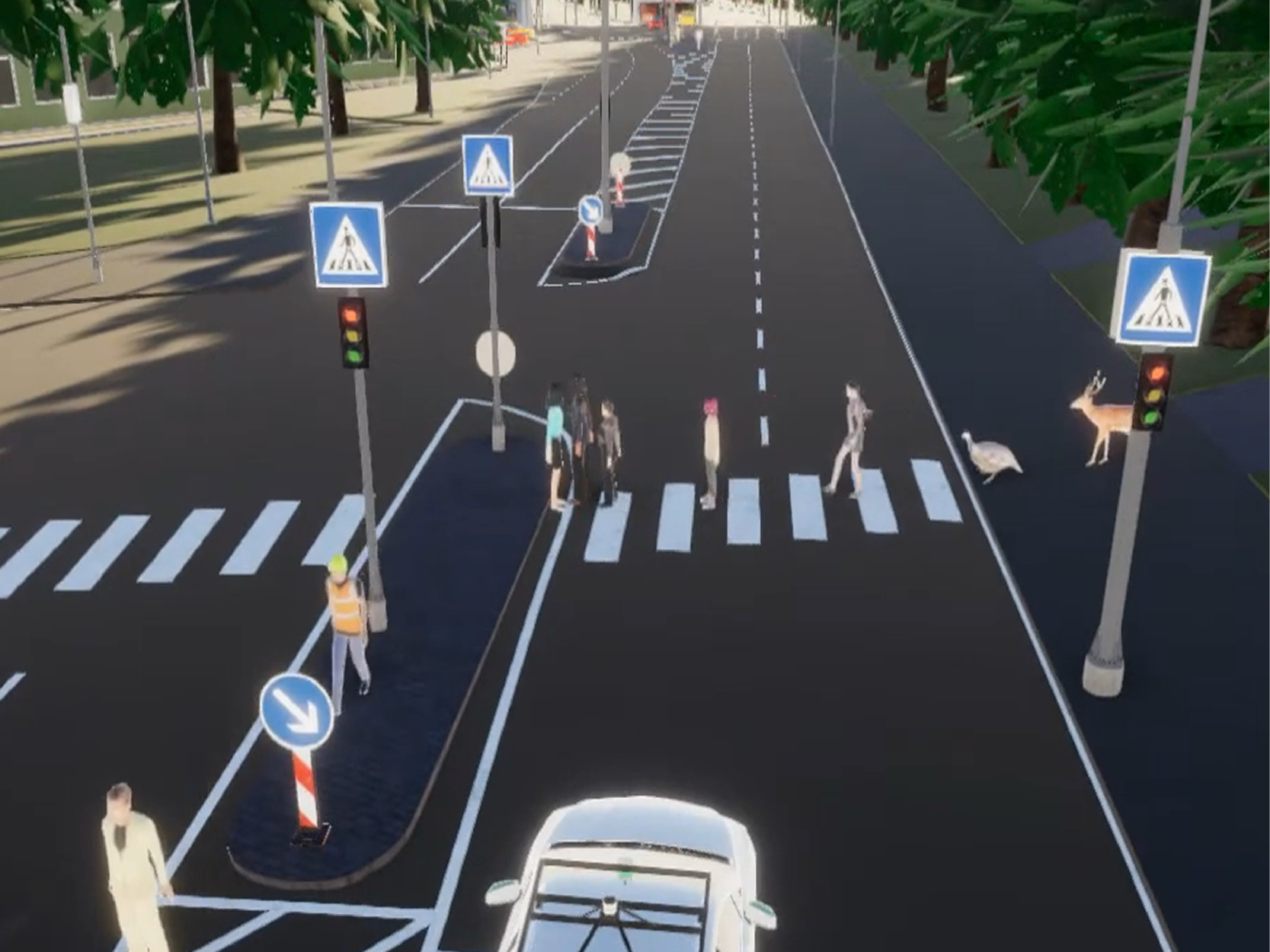} \\
        \textbf{LGSVL}\\
        \raggedright \footnotesize Limited assets, no skeleton-based control
    \end{minipage}
    \caption{Overview of existing simulators highlighting limitations in pedestrian representation.}
    \label{fig:comparison}
\end{figure}

Figure~\ref{fig:comparison} provides a comparative overview of three widely used simulation platforms—AirSim, Gazebo, and LGSVL—highlighting their respective limitations in representing pedestrians. Accurately modeling pedestrian behavior, especially rare or unconventional gestures, is critical for evaluating autonomous driving systems in realistic traffic scenarios.  

\textbf{AirSim}~\cite{airsim2017fsr} is an open-source simulator developed by Microsoft on top of Unreal Engine, primarily targeting autonomous vehicle and drone research. It offers high-fidelity visual and physical simulation, making it well-suited for environmental realism and sensor testing. However, AirSim lacks a native pedestrian asset class, limiting the ability to simulate complex pedestrian behaviors or interactions. Consequently, scenarios involving vulnerable road users cannot be fully represented, reducing the applicability of AirSim for comprehensive autonomous driving evaluation.  

\textbf{Gazebo}~\cite{xie2023drl} is a versatile robotics simulator widely used in conjunction with the Robot Operating System (ROS). It incorporates a social-force model to simulate pedestrian movement and crowd dynamics, providing a basic framework for multi-agent interactions. Despite this, Gazebo does not support gestures or skeleton-based control, meaning pedestrians cannot perform nuanced behaviors such as hand signals, sudden evasive actions, or other uncommon motions. This restricts its utility in testing autonomous agents under diverse human behaviors.  

\textbf{LGSVL}~\cite{rong2020lgsvl} offers realistic sensor modeling and urban traffic scenarios. While it includes pedestrian agents, these are limited in number and diversity, and the platform does not provide skeleton-based control for complex motion generation. Pedestrian movements are typically predefined along fixed paths, preventing the simulation of spontaneous or interactive behaviors that autonomous agents are likely to encounter in real-world conditions.  

Collectively, these limitations illustrate a critical gap in existing simulation platforms: the inability to realistically model pedestrian behaviors, particularly rare or unconventional gestures. This gap motivates the design of the HABIT benchmark, which explicitly incorporates behaviorally diverse pedestrians capable of performing a wide range of motions and gestures. HABIT aims to rigorously evaluate autonomous agents under realistic and challenging traffic interactions. This ensures a more comprehensive assessment of agent competence in environments that closely mimic real-world complexities.

\begin{figure}[!h]
    \centering
    \begin{minipage}[t]{0.32\linewidth}
        \centering
        \includegraphics[width=\linewidth]{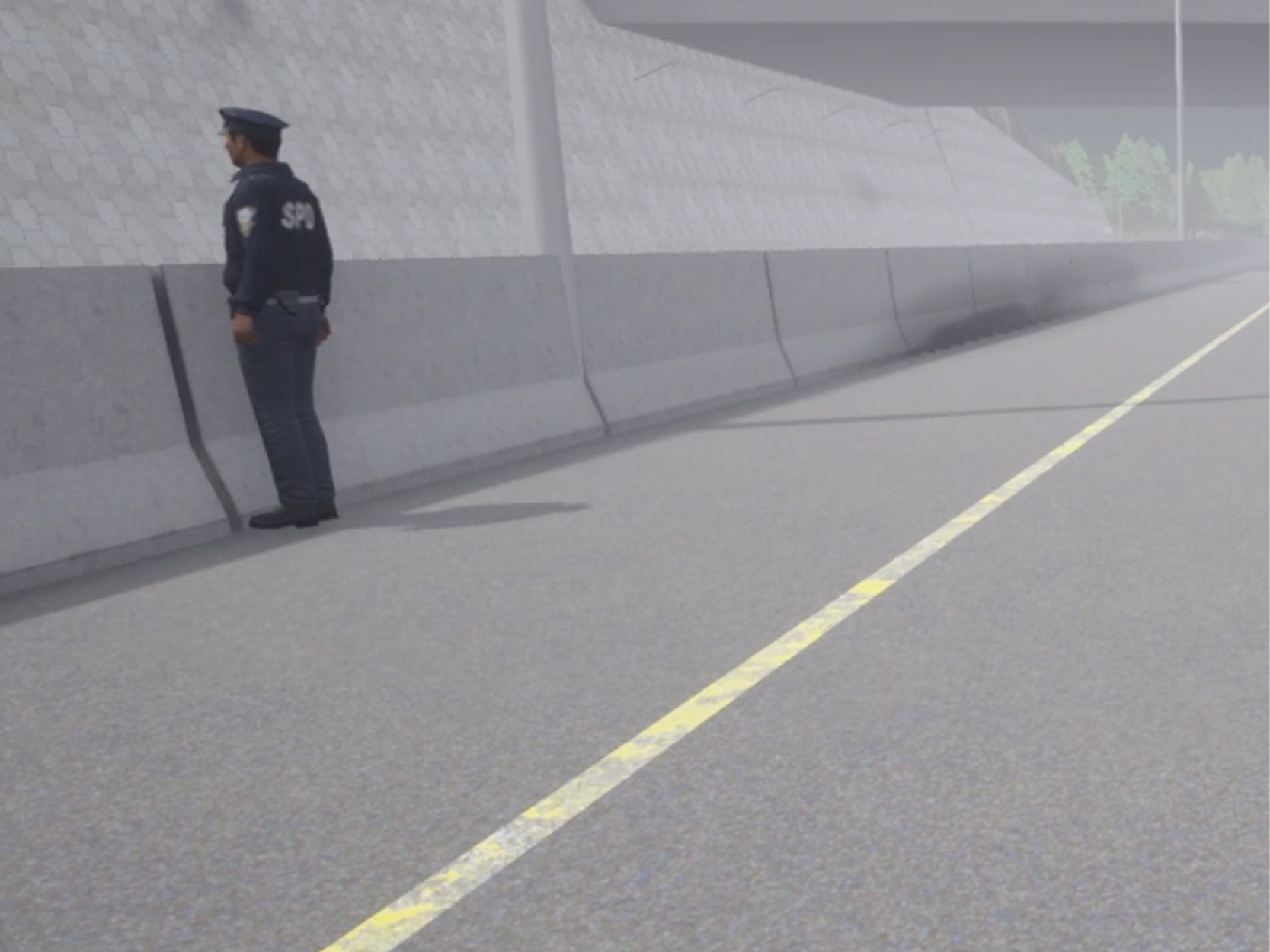} \\
        % \textbf{Airsim}~\cite{airsim2017fsr} \\
        % \raggedright \footnotesize No native pedestrian asset class
    \end{minipage}%
    \hfill
    \begin{minipage}[t]{0.32\linewidth}
        \centering
        \includegraphics[width=\linewidth]{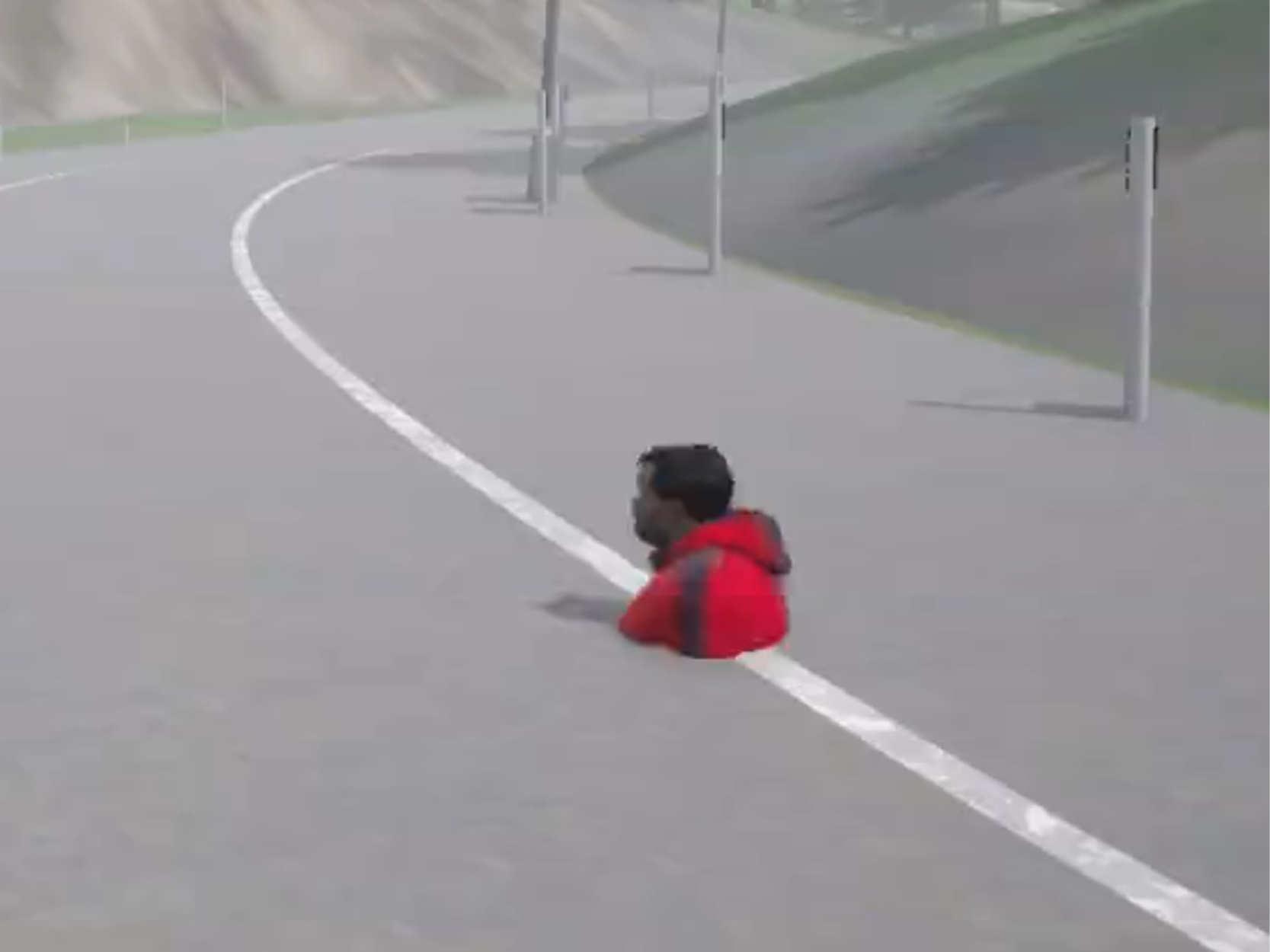} \\
        % \textbf{Gazebo}~\cite{xie2023drl} \\
        % \raggedright \footnotesize Social-force pedestrian model, no gestures
    \end{minipage}%
    \hfill
    \begin{minipage}[t]{0.32\linewidth}
        \centering
        \includegraphics[width=\linewidth]{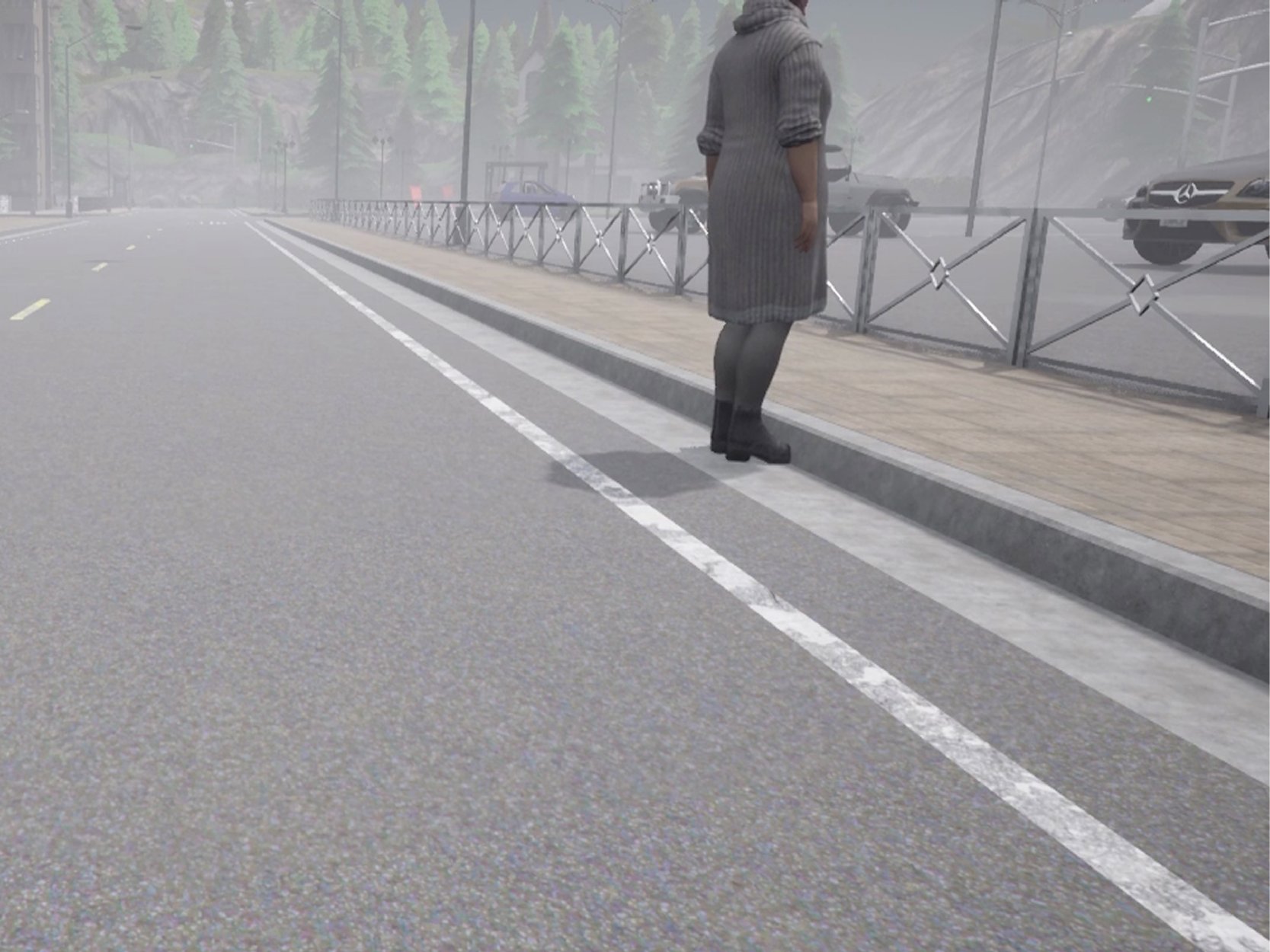} \\
        % \textbf{LGSVL}~\cite{rong2020lgsvl} \\
        % \raggedright \footnotesize Limited assets, no skeleton-based control
    \end{minipage}
    \caption{Examples from the ARCANE project (CARLA-BSP): scenes featuring a single pedestrian in crossing/non-crossing situations, following binary or simplified behavior labels.}
    \label{fig:arcane}
\end{figure}

CARLA, by contrast, provides skeleton (bone) control of pedestrian models, enabling far finer‐grained articulation and behavioral richness than many other simulators. However, widespread use of this capability remains unrealized owing to a variety of complications: retargeting motion capture or motion data to CARLA’s coordinate systems, resolving mismatches in animation rigs and skeleton hierarchies, and implementing suitable collision, trajectory, and motion blending to avoid artifacts. One recent project, ARCANE~\cite{wielgosz2023carlabsp}, used CARLA’s pedestrian catalogue for pedestrian intention estimation. Still, this work was constrained: pedestrians only followed pre-defined paths (without dynamic backtracking or deviations), demonstrated minimal interaction with the environment (e.g. no directional changes, no collision avoidance), and often suffered from simulation physics anomalies (e.g. pedestrians ‘tunneling’ through geometry or going underground) or lack of background traffic context. A few such failure modes are shown in Figure~\ref{fig:arcane}

\begin{figure}[!h]
    \centering
    \begin{minipage}[t]{0.32\linewidth}
        \centering
        \includegraphics[width=\linewidth]{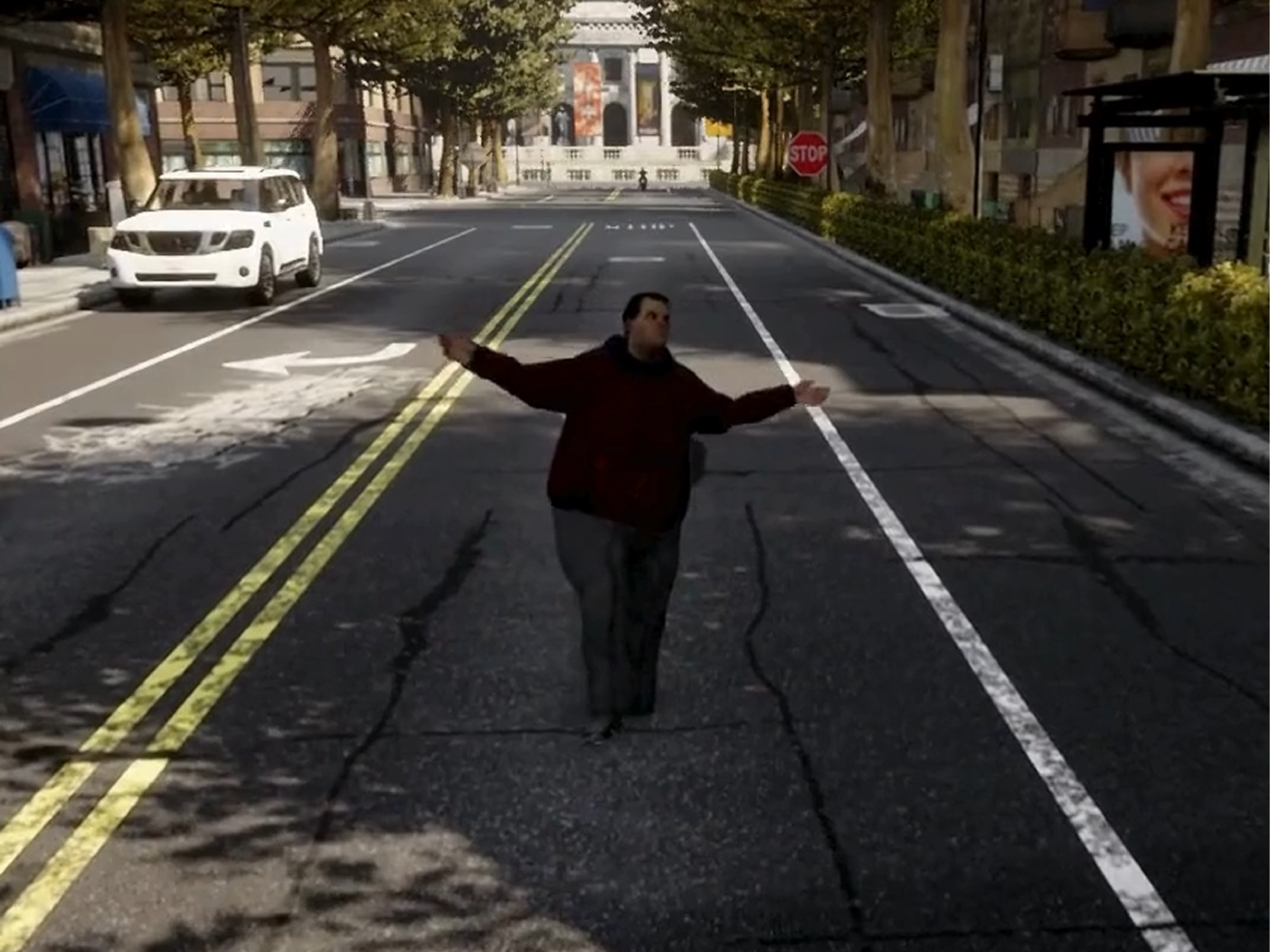} \\
        % \textbf{Airsim}~\cite{airsim2017fsr} \\
        % \raggedright \footnotesize No native pedestrian asset class
    \end{minipage}%
    \hfill
    \begin{minipage}[t]{0.32\linewidth}
        \centering
        \includegraphics[width=\linewidth]{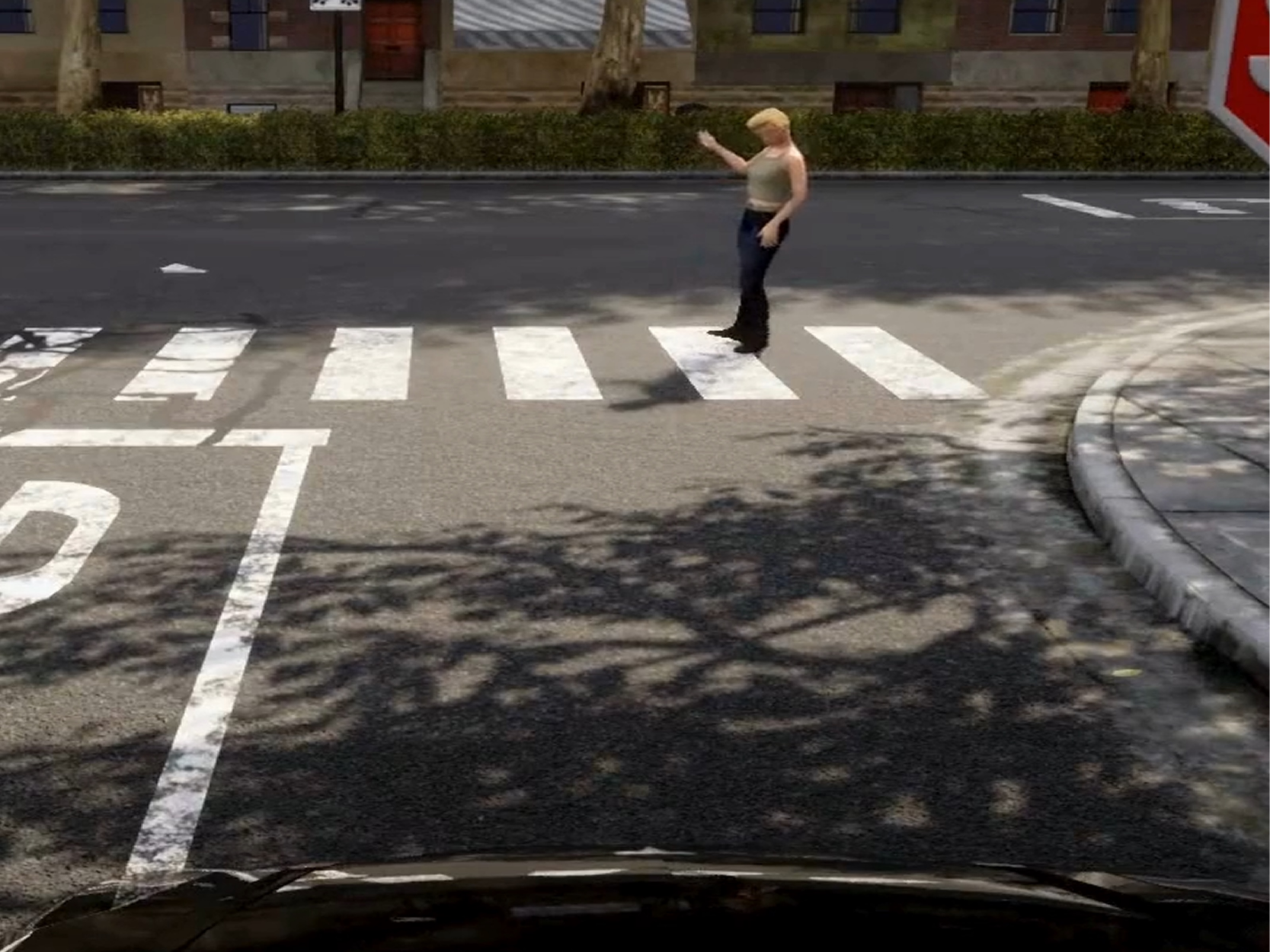} \\
        % \textbf{Gazebo}~\cite{xie2023drl} \\
        % \raggedright \footnotesize Social-force pedestrian model, no gestures
    \end{minipage}%
    \hfill
    \begin{minipage}[t]{0.32\linewidth}
        \centering
        \includegraphics[width=\linewidth]{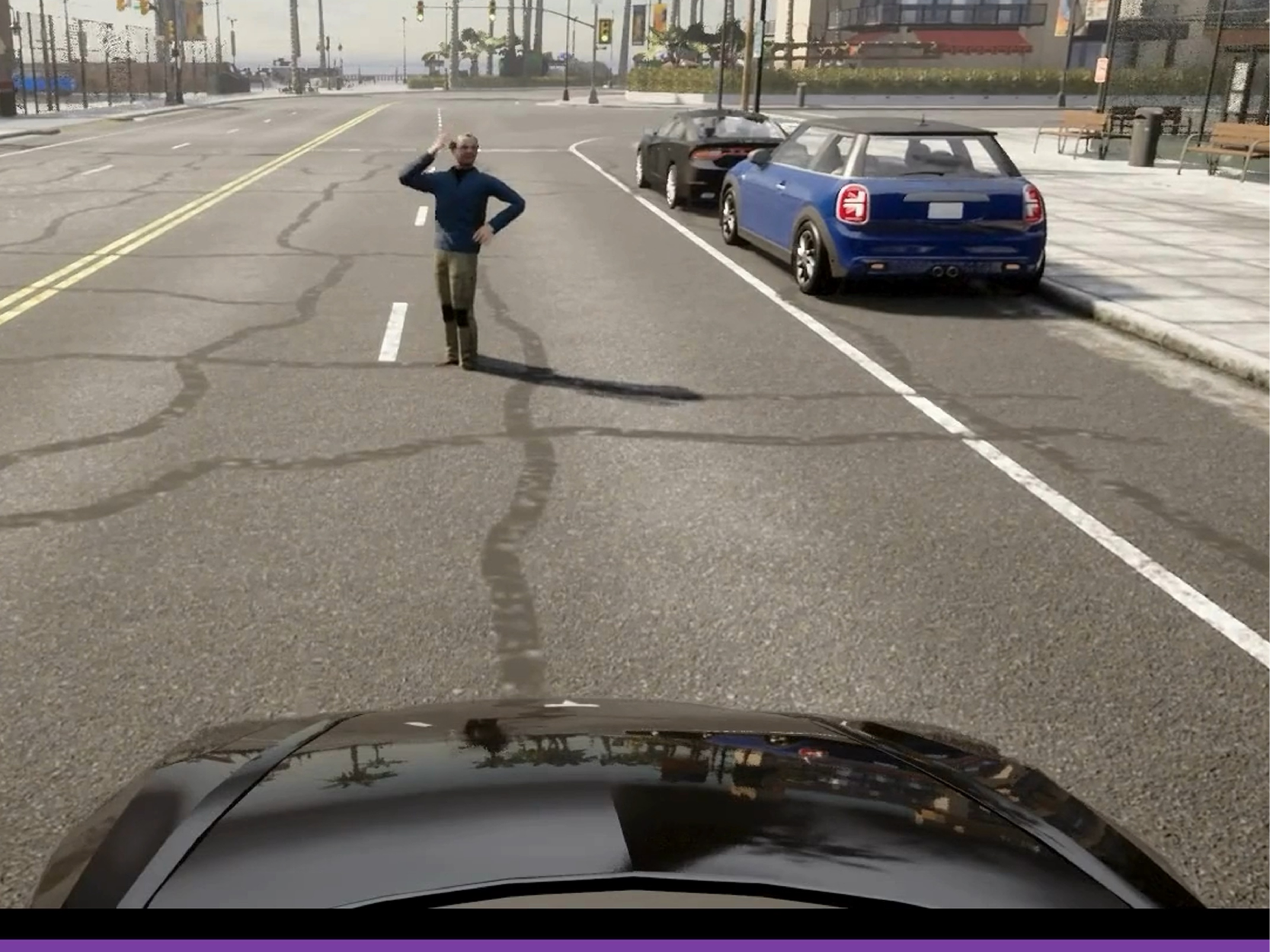} \\
        % \textbf{LGSVL}~\cite{rong2020lgsvl} \\
        % \raggedright \footnotesize Limited assets, no skeleton-based control
    \end{minipage}
    \caption{Examples of rare pedestrian behaviours realized through the HABIT framework, demonstrating gestures, non-linear trajectories, interaction with environment, and reactive adaptation.}
    \label{fig:habit_rare}
\end{figure}

Our HABIT framework is engineered to overcome these limitations. We systematically retarget open motion sources to CARLA’s pedestrian asset skeletons. By integrating velocity grounding, trajectory reconstruction, and collision avoidance, we convert general-purpose motion data into agents who behave as active traffic participants. In Figure~\ref{fig:habit_rare} we present hand-crafted instances of rare or atypical pedestrian behaviour that are currently only possible with our approach.

\section{Retargeting Fidelity}

The motion transfer is fundamentally deterministic, with fidelity limited primarily by the mathematical constraints of Euler angle representation (gimbal lock) rather than algorithmic approximation. Extensive ground-truth comparison with SMPL data would not address the fundamental mathematical constraint we’ve isolated and solved.

\begin{figure}[!h]
    \centering
    \begin{minipage}[t]{0.32\linewidth}
        \centering
        \includegraphics[width=\linewidth]{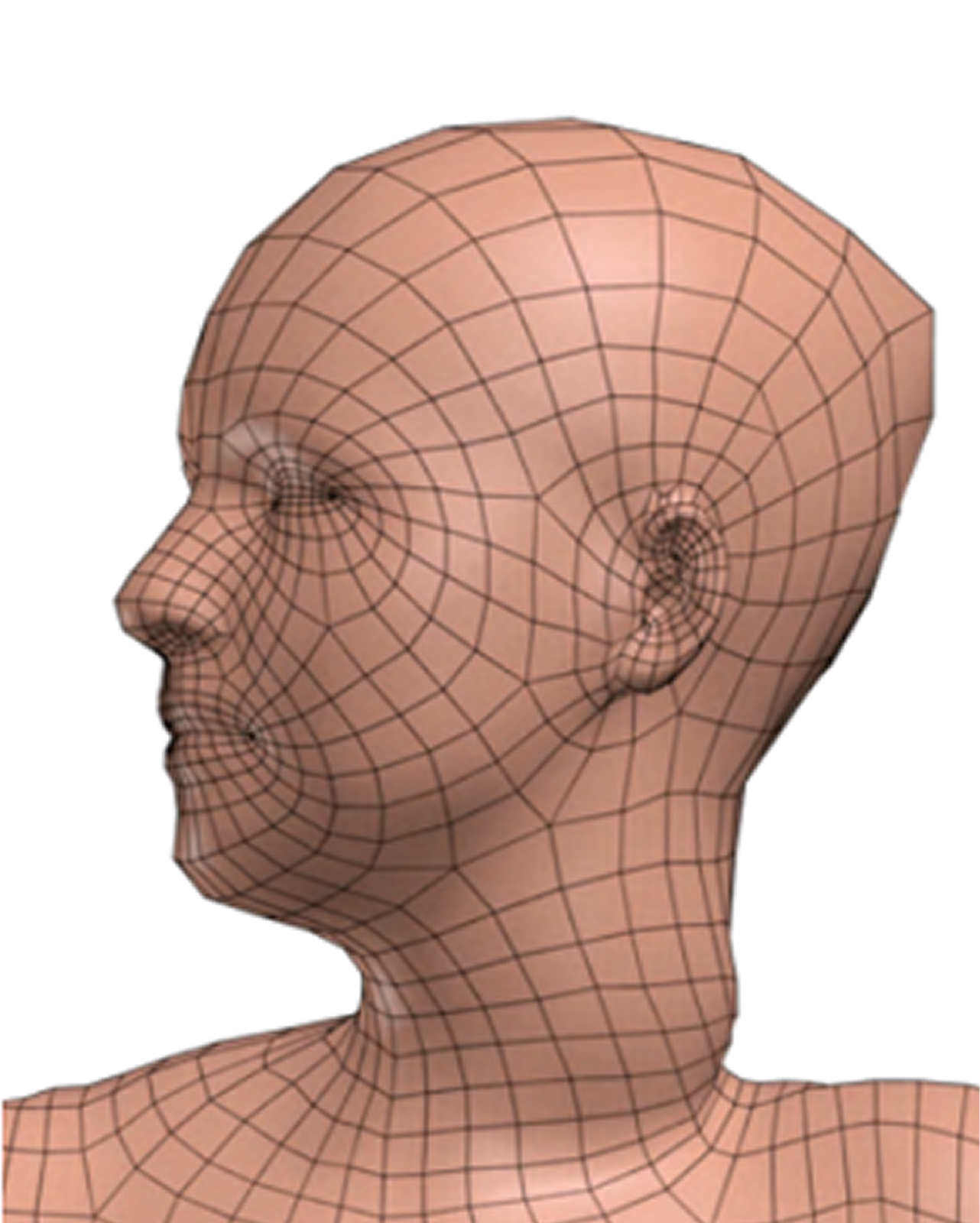} \\
        \footnotesize Euler angles
    \end{minipage}%
    \hspace{0.05\linewidth} % small horizontal space between images
    \begin{minipage}[t]{0.32\linewidth}
        \centering
        \includegraphics[width=\linewidth]{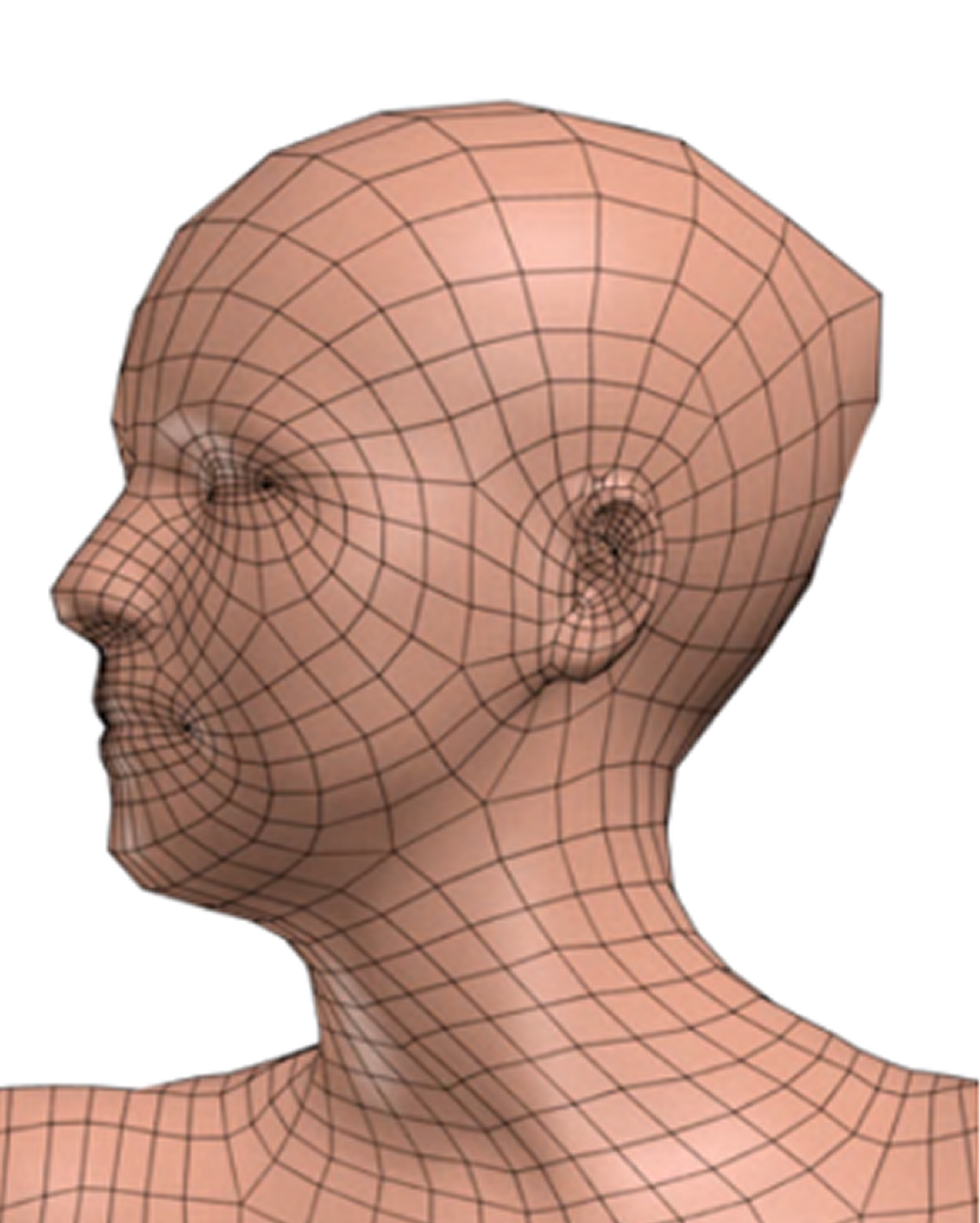} \\
        \footnotesize SMPL parameterization
    \end{minipage}
    \caption{Pose blend shape deformation due to Euler angle approximation of SMPL data.}
    \label{fig:habit_rare}
\end{figure}

\begin{figure}[!h]
  \centering
  \includegraphics[width=0.75\linewidth, height=0.35\linewidth]{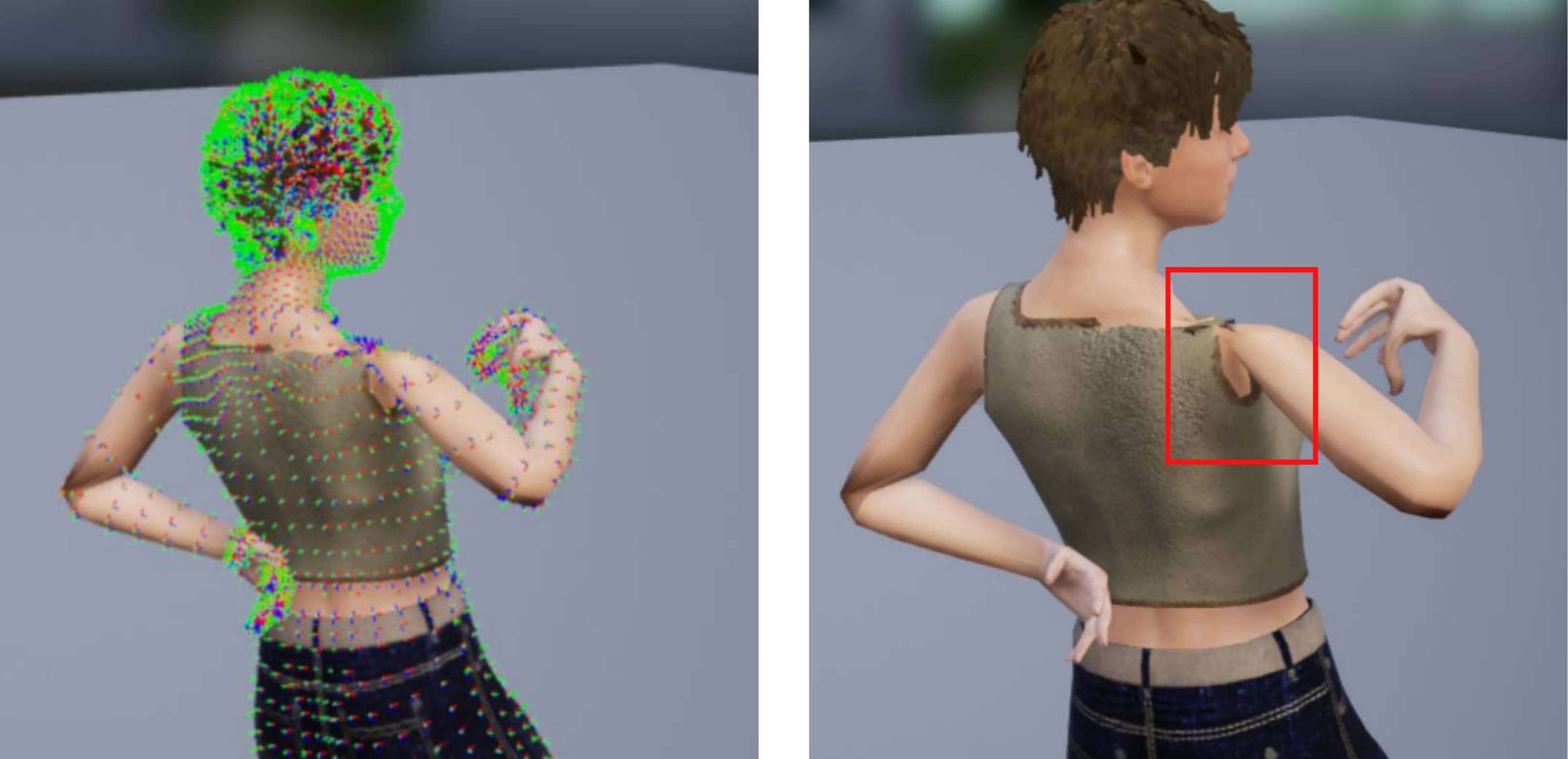}
  \caption{
    \textbf{Retargeting fidelity} 
       At extreme positions the character mesh deforms as a native limitation of Euler angle approximation, however the transfered pose remains skeletally same.
    }
  \label{fig:fidelity}
\end{figure}

\section{Perception realism of HABIT}

Zero-shot perception tasks offer a robust method for evaluating visual realism and domain fidelity in simulated environments, particularly when using models trained solely on real-world data. Successful zero-shot performance in detection, segmentation, tracking, and pose estimation tasks strongly indicates alignment with real-world visual and contextual distributions \cite{Peng18,Varol17,Andriluka18}. Specifically, human-centric perception tasks implicitly validate realism through accurate modeling of human shape, articulation, and environmental interaction.

\begin{figure}[htbp]
  \centering
  \includegraphics[width=\linewidth]{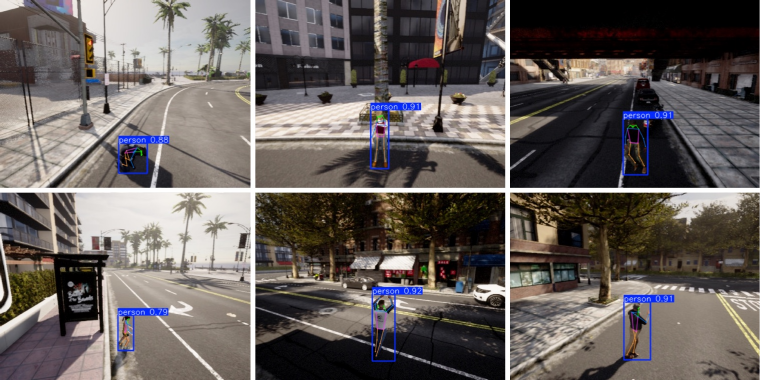}
  \caption{
    \textbf{Zero-shot pose estimation on simulated data.} Qualitative results of YOLOv11l-pose applied to HABIT.
  }
  \label{fig:yolo}
\end{figure}

We perform zero-shot evaluation using \textit{YOLOv11l-pose}, a keypoint-aware variant of YOLOv11 trained exclusively on real-world data~\cite{Lin14,Cao17}, on 12,000 randomly sampled HABIT frames (5 per motion sequence). Results in Table~\ref{tab:pose_metrics} confirm high performance, achieving \textbf{$93.4\%$} mAP\text{@}$0.5$ for detection and \textbf{$95.7\%$} PCK\text{@}$0.5$ for pose estimation, closely aligning with real-world results using various YOLOv11 models~\cite{Varol17,Peng18}. These findings validate HABIT’s visual and biomechanical fidelity and its utility for benchmarking pedestrian-aware systems~\cite{hu2023simulation}.

\begin{table}[h]
  \centering
  \resizebox{\linewidth}{!}{
    \begin{tabular}{@{}lcc@{}}
      \toprule
      \textbf{Metric} & \textbf{HABIT (Synthetic)} & \textbf{Real-World (COCO)} \\
      \midrule
      PCK\text{@}0.2            & $0.692 \pm 0.223$  & $0.65$–$0.75$ \\
      PCK\text{@}0.5            & $0.957 \pm 0.144$  & $>0.95$ \\
      MPJPE (px)         & $9.3 \pm 9.6$      & $8$–$12$ \\
      OKS                & $0.418 \pm 0.169$  & $0.40$–$0.55$ \\
      mAP@[.5:.95]       & $0.586 \pm 0.269$  & $0.55$–$0.65$ \\
      mAP@0.5            & $0.934 \pm 0.248$  & $>0.90$ \\
      IoU (bbox overlap) & $0.766 \pm 0.164$  & $0.75$–$0.80$ \\
      \bottomrule
    \end{tabular}
  }
  \caption{Comparison of pose estimation performance on the HABIT and real-world COCO datasets. Similar trends highlight the visual realism and fidelity of the HABIT benchmark.}
  \label{tab:pose_metrics}
\end{table}

\begin{figure}[htbp]
  \centering
  \includegraphics[width=\linewidth, keepaspectratio]{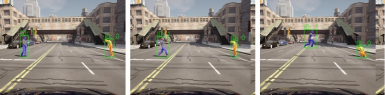}
  \caption{
    \textbf{Multi-frame detection, segmentation, and tracking.} 
        Consistent track IDs, even in edge-case motions, highlight pose consistency and temporal coherence of HABIT.
    }
  \label{fig:tracking}
\end{figure}

To assess spatial consistency and trackability, we apply the Segment Anything Model (SAM)~\cite{kirillov2023segment} to YOLO-pose detections. Using YOLO-derived keypoints for guidance, we generate segmentation masks and track identities across frames. Qualitative results ( Figure~\ref{fig:tracking}) demonstrate accurate and temporally consistent segmentation, despite SAM being trained solely on real-world data, highlighting HABIT’s visual continuity and its suitability for multi-object tracking without domain-specific adaptation."

We further assess the realism of HABIT pedestrians through segmentation quality using the Segment Anything Model (SAM) trained solely on real-world data. This provides a zero-shot measure of how well simulated pedestrians align with real distributions. \textbf{We report two standard metrics:}

\begin{itemize}
    \item Intersection over Union (IoU): overlap between predicted and reference masks.
    \item Dice Coefficient (DICE): similarity measure that is more sensitive to boundary alignment.
\end{itemize}

\begin{table}[htbp]
\centering
\caption{SAM Evaluation Results on Pedestrians of CARLA}
\label{tab:sam_evaluation}
\begin{tabular}{lccc}
\hline
\textbf{Metric} & \textbf{Mean ± Std} & \textbf{Median} & \textbf{90th Percentile} \\
\hline
IoU Score & 0.7364 ± 0.3333 & 0.8974 & 0.9392 \\
DICE Score & 0.7824 ± 0.3443 & 0.9459 & -- \\
\hline
\multicolumn{4}{l}{\textit{Pipeline success rate: 95.9\%}} \\
\hline
\end{tabular}
\end{table}

Median IoU (0.90) and Dice (0.95) indicate high segmentation fidelity, with over 70\% of cases achieving excellent quality. These results confirm that HABIT pedestrians exhibit strong visual and structural realism suitable for benchmarking perception models.
%-------------------------------------------------------------------------

% The reviewer ids are mapped as follows
% \textcolor{r1color}{R1}: ievy \textcolor{r2color}{R2}: WZEa \textcolor{r3color}{R3}: 9dd3 \textcolor{r4color}{R4}: 5BqF

\end{document}